\documentclass[]{interact}

\usepackage{newtxtext,newtxmath} 
\usepackage{tikz}
\usepackage{tikz-3dplot}
\usepackage{amsfonts} 
\usepackage{amsmath,amsthm}
\usepackage[mathscr]{euscript}
\usepackage[noend]{algpseudocode}
\usepackage{booktabs}
\usepackage{algorithm}
\usepackage{natbib}
\usepackage{enumitem}
\usepackage{hyperref}
\usepackage{multirow}
\usepackage{caption}
\usepackage{subcaption}
\usepackage{xcolor}
\usepackage{graphics}

\usepackage{figure_shapes}

\def\R{\mathbb{R}}
\newcommand{\ts}[1]{\pmb{\mathscr{#1}}}
\newcommand{\kruskal}[1]{\pmb{\mathfrak{#1}}}
\newcommand{\nspace}[1]{\pmb{\mathcal{#1}}}

\theoremstyle{plain}
\newtheorem{theorem}{Theorem}[section]

\newtheorem{proposition}[theorem]{Proposition}

\theoremstyle{definition}

\theoremstyle{remark}

\begin{document}


\title{Coupled Support Tensor Machine Classification for Multimodal Neuroimaging Data}

\author{ Li Peide*, Seyyid Emre Sofuoglu*\footnote{* Equal contribution.}, Tapabrata Maiti, Selin Aviyente
}

\maketitle

\begin{abstract}
Multimodal data arise in various applications where information about the same phenomenon is acquired from multiple sensors and across different imaging modalities. Learning from multimodal data is of great interest in machine learning and statistics research as this offers the possibility of capturing complementary information among modalities. Multimodal modeling helps to explain the interdependence between heterogeneous data sources, discovers new insights that may not be available from a single modality, and improves decision-making.  Recently, coupled matrix-tensor factorization has been introduced for multimodal data fusion to jointly estimate latent factors and identify complex interdependence among the latent factors. However, most of the prior work on coupled matrix-tensor factors focuses on unsupervised learning and there is little work on supervised learning using the jointly estimated latent factors. This paper considers the multimodal tensor data classification problem. A Coupled Support Tensor Machine (C-STM) built upon the latent factors jointly estimated from the Advanced Coupled Matrix Tensor Factorization (ACMTF) is proposed. C-STM combines individual and shared latent factors with multiple kernels and estimates a maximal-margin classifier for coupled matrix tensor data. The classification risk of C-STM is shown to converge to the optimal Bayes risk, making it a statistically consistent rule. C-STM is validated through simulation studies as well as a simultaneous EEG-fMRI analysis. The empirical evidence shows that C-STM can utilize information from multiple sources and provide a better classification performance than traditional single-mode classifiers.

\end{abstract}


\section{Introduction}
Advances in clinical neuroimaging and computational bioinformatics have dramatically increased our understanding of various brain functions using multiple modalities such as Magnetic Resonance Imaging (MRI), functional Magnetic Resonance Imaging (fMRI), electroencephalogram (EEG), and Positron Emission Tomography (PET). Their strong connections to the patients' biological status and disease pathology suggest the great potential of their predictive power in disease diagnostics. Numerous studies using vector- and tensor-based statistical models illustrate how to utilize these imaging data both at the voxel- and Region-of-Interest (ROI) level and develop efficient biomarkers that predict disease status. For example, \citet{anderson2011functional} proposes a classification model using functional connectivity MRI for autism disease and reaches 89\% diagnostic accuracy for subjects under 20. \citet{schindlbeck2018network} utilizes network models and brain imaging data to develop novel biomarkers for Parkinson's disease. Many works in Alzheimer's disease research such as \citet{morris2009pittsburgh, liu2014early, khazaee2016application, gavidia2017early, long2017prediction,ding2019deep, li2020early} use EEG, MRI and PET imaging data to predict patient's cognition and detect early-stage Alzheimer's diseases. Although these studies have provided impressive results, utilizing imaging data from a single modality such as individual MRI sequences are known to have limited predictive capacity, especially in the early phases of the disease. For instance, \citet{li2020early} uses brain MRI volumes from regions of interest to identify patients in early-stage Alzheimer's disease. They use one-year MRI data from Alzheimer’s Disease Neuroimaging Initiative (ADNI) and obtain 77\% prediction accuracy. Although such a performance is favorable compared to other existing approaches, the diagnostic accuracy is relatively low due to the limited information from MRI data. In recent years, it has been common to acquire multiple neuroimaging modalities in clinical studies such as simultaneous EEG-fMRI or MRI and fMRI. Even though each modality measures different biological signals, they are interdependent and mutually informative.
Learning from multimodal neuroimaging data may help integrate information from multiple sources and facilitate biomarker developments in clinical studies. It also raises the need for novel supervised learning techniques for multimodal data in statistical learning literature. 

Existing statistical approaches to multimodal data science are dominated by unsupervised learning methods. These methods analyze multimodal neuroimaging data by performing joint matrix decomposition and extracting common information across different modalities. During optimization, the decomposed factors bridging two or more modalities are estimated to interpret the connections between different modalities. Examples of these methods include matrix-based joint Independent Component Analysis (jICA) \citet{calhoun2006method, groves2011linked, lei2012eeg, liu2009combining, sui2011discriminating, acar2019unraveling} which assume bilinear correlations between factors in different modalities. However, these matrix-vector based models cannot preserve the multilinear nature of original data and the spatiotemporal correlations across modes as most neuroimaging modalities are naturally in tensor format. Recently, various coupled matrix-tensor decomposition methods have been introduced to address this issue  \citet{acar2014structure, acar2017tensor, acar2019unraveling, chatzichristos2018fusion, chatzichristos2020early, karahan2015tensor, mosayebi2020correlated}. These methods impose different soft or hard multilinear constraints between factors from different modalities providing more flexibility in data modeling. 

Current supervised learning approaches for multimodal data mostly concatenate data modalities as extra features without exploring their interdependence. For example, \citet{zhou2013tensor, li2018tucker} build generalized regression models by appending tensor and vector predictors linearly for image prediction and classification. \citet{pan2018covariate} develops a discriminant analysis  by including tensor and vector predictors in a linear fashion. \citet{li2019integrative} proposes an integrative factor regression for multimodal neuroimaging data assuming that data from different modalities can be decomposed into latent factors. More recently, \citet{gahrooei2021multiple} proposed multiple tensor-on-tensor regression for multimodal data, which combines tensor-on-tensor regression from \citet{lock2018tensor} with traditional additive linear model. Another type of integration utilizes kernel tricks and combines information from multimodal data with multiple kernels. \citet{gonen2011multiple} provides a survey on various multiple kernel learning techniques for multimodal data fusion and classification with support vector machines. Combining kernels linearly or non-linearly instead of original data in different modalities provides more flexibility in information integration. \citet{bach2008consistency} proposed a multiple kernel regression model with group lasso penalty, which integrates information by multiple kernels and selects the most predictive data modalities. 

Despite these accomplishments, the current approaches have several shortcomings. First, they mainly focus on exploring the interdependence between multimodal imaging data, ignoring the representative and discriminative power of the learned components. Thus, the methods cannot further bridge the imaging data to the patients' biological status, which is not helpful in biomarker development. Second, the supervised techniques such as  integrate information primarily by data or feature concatenation without explicitly considering the possible correlations between different modalities. This lack of consideration of interdependence may cause issues like overfitting and parameter identifiability. Third, even though methods from \citet{li2019integrative, gahrooei2021multiple} have considered latent structures for multimodal data, these models are designed primarily for linear regression and are not directly applicable to classification problems. Fourth, the aforementioned multimodal analysis methods are mainly vector based methods, which cannot handle large-size multi-dimensional data encountered in contemporary data science. As discussed in \citet{bi2020tensors}, tensors provide a powerful tool for analyzing multi-dimensional data in statistics. As a result, developing a novel multimodal tensor-based statistical framework for supervised learning can be of great interest. Finally, although many empirical studies demonstrate the success of using multimodal data, there is a lack of mathematical and statistical clarity to the extent of generalizability and associated uncertainties. The absence of a solid statistical framework for multimodal data analysis makes it impossible to interpret the generalization ability of a certain statistical model. 

In this paper, we propose a two-stage Coupled Support Tensor Machine (C-STM) for multimodal tensor-based neuroimaging data classification. The proposed model addresses the current issues in multimodal data science and provides a sound statistical framework to interpret the interdependence between modalities and quantify the model consistency and generalization ability. The \textbf{major contributions} of this work are:
\begin{enumerate}
    \item Individual and common latent factors are extracted from multimodal tensor data, for each sample or subject, using Advanced Coupled Matrix Tensor Factorization (ACMTF) \citep{acar2014structure, acar2017acmtf}. The extracted components are then utilized in a statistical framework. Most of the work on ACMTF do not work on each subject separately and the extracted factors are utilized for a signal analysis rather than a subsequent statistical learning framework. Specifically, the work on supervised approaches with CMTF is limited.
    
    \item Building a novel Coupled Support Tensor Machine with both the coupled and non-coupled tensor CP factors for classification. In this regard, multiple kernel learning approaches are adopted to integrate components from multi-modal data.
    
    \item For the validation of our work, we provide both theoretical and empirical evidence. We provide theoretical results such as classification consistency for statistical guarantee. A thorough numerical study has been conducted, including a simulation study and experiments on real data to illustrate the usefulness of the proposed methodology.

\end{enumerate}
A Matlab package is also provided in the supplemental material, including all functions for C-STM classification. The source codes are available at our Github repository \footnote{\url{https://github.com/mrsfgl/supervised-tensor-fusion}}.

\section{Related Work}
\label{sec:related_}
In this section, we review some background and prior work on tensor decompositions and multiple kernel learning.

\subsection{Notations}
\label{sec:tensoralgebra_chap1}
In this work, we denote numbers and scalars by letters such as $x, y, N$. Vectors are denoted by boldface lowercase letters, e.g. $\pmb{a}, \pmb{b}$. Matrices are denoted by boldface capital letters like $\pmb{A}, \pmb{B}$. Multi-dimensional tensors are denoted by boldface Euler script letters such as $\ts{X}, \ts{Y}$. The order of a tensor is the number of dimensions of the data hypercube, also known as ways or modes. For example, a scalar can be regarded as a zeroth-order tensor, a vector is a first-order tensor, and a matrix is a second-order tensor. 

Let $\ts{X}\in\R^{I_1\times I_2\times \dots \times I_N}$ be a tensor of order $N$, where $x_{i_1,i_2,\dots,i_N}$ denotes the $({i_1,i_2,\dots,i_N})$th element of the tensor. Vectors obtained by fixing all indices of the tensor except the one that corresponds to $n$th mode are called mode-$n$ fibers and denoted as $\pmb{x}_{i_1,\dots i_{n-1},i_{n+1},\dots i_N}\in\R^{I_n}$. The mode-$n$ unfolding of $\ts{X}$ is defined as $\ts{X}_{(n)}\in \R^{I_n\times \prod_{n'=1,n'\neq n}^NI_{n'}}$ where the mode-$n$ fibers of the tensor $\ts{X}$ are the columns of $\ts{X}_{(n)}$ and the remaining modes are organized accordingly along the rows.

\subsection{Canonical/Polyadic (CP) Decomposition}
Let $\ts{X} \in \mathbb{R}^{I_{1} \times I_2 \times \ldots \times I_{d}}$ be a tensor with $d$ modes. Rank-$r$ Canonical/Polyadic (CP) decomposition of $\ts{X}$ is defined as:
\begin{equation}
\begin{split}
        \ts{X} &\approx \sum \limits_{k = 1}^{r}  \bm{x}_{k}^{(1)} \circ \bm{x}_{k}^{(2)} ... \circ \bm{x}_{k}^{(d)} = \llbracket \bm{X}^{(1)},..., \bm{X}^{(d)} \rrbracket
\end{split}
\label{equ:Trank-r}
\end{equation}
where $\bm{X}^{(j)} \in \mathbb{R}^{I_{j}\times r}, j = 1,..,d$ are defined as factor matrices whose columns are $\pmb{x}_{r}^{(j)}$ and "$\circ$" represents the vector outer product. The second equality in (\ref{equ:Trank-r}) is called Kruskal tensor (see \citet{Kruskal_Three_1977}), which is a convenient representation for CP tensors. We denote a Kruskal tensor by $\kruskal{U}_{x} = \llbracket \bm{X}^{(1)},..., \bm{X}^{(d)}\rrbracket$ or $\kruskal{U}_x=\llbracket \pmb{\zeta};\bm{X}^{(1)},..., \bm{X}^{(d)}\rrbracket$ where $\pmb{\zeta}\in\R^{r}$ is a vector whose entries are the weights of rank one tensor components. In the special case of matrices, $\pmb{\zeta}$ corresponds to singular values of a matrix. In general, it is assumed that the rank $r$ is small so that equation (\ref{equ:Trank-r}) is also called low-rank approximation for a tensor $\ts{X}$. Such an approximation can be estimated by an Alternating Least Square (ALS) approach (see \citet{kolda2009tensor}).

Although there are other tensor decomposition structures such as Tucker Decomposition or Tensor Train Decomposition, the advantage of CP is that the factors extracted by a CP decomposition are unique up to permutation. The uniqueness of the factors make CP decomposition more interpretable. 

\subsection{Coupled Matrix Tensor Factorization}
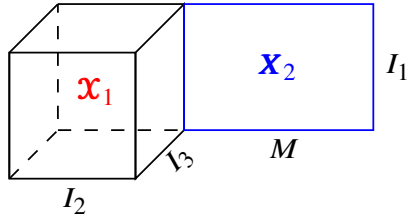
\begin{figure}
    \centering
    \resizebox{0.4\textwidth}{!}{
    \begin{tikzpicture}
\coordinate (O) at (0 , 0, 0);
\coordinate (A) at (0, 1, 0);
\coordinate (B) at (0, 1, 1);
\coordinate (C) at (0, 0, 1);
\coordinate (D) at (1, 1, 1);
\coordinate (E) at (1, 1, 0);
\coordinate (F) at (1, 0, 0);
\coordinate (G) at (1, 0, 1);

\draw[black] (A) -- (B) -- (C);
\draw[black] (A) -- (E);
\draw[dashed] (C) --(O) -- (A);
\draw[dashed] (O) -- (F);
\draw[black] (C) -- (B) -- (D) -- (G) -- cycle;
\draw[black] (G) -- (D) -- (E) -- (F) -- cycle;

\draw (0.5, 0.5, 0.5) node[rotate = 0] {{\tiny \color{red} $\ts{X}_{1}$}};
\draw (0.6, -0.1, 1.2) node[rotate = 0] {{\tiny $I_{2}$}};
\draw (1.2, 0, 0.6) node[rotate = 45] {{\tiny $I_{3}$}};
\draw (2.7, 0.5, 0) node[rotate = 0] {{\tiny $I_{1}$}};
\draw (1.8, -0.15, 0) node[rotate = 0] {{\tiny $M$}};
\draw (1.75, 0.5, 0) node[rotate = 0] {{\tiny \color{blue} $\pmb{X}_{2}$}};

\coordinate (O) at (1 , 0, 0);
\coordinate (A) at (2.5, 0, 0);
\coordinate (B) at (2.5, 1, 0);
\coordinate (C) at (1, 1, 0);

\draw[blue] (O) -- (A) -- (B) -- (C) -- cycle;
\end{tikzpicture}}
    \caption{Illustration of Coupled Tensor Matrix Model}
    \label{fig:data_couple_}
\end{figure}

Motivated by the fact that joint analysis of data from multiple sources can potentially unveil complex data structures and provide more information, Coupled Matrix Tensor Factorization (CMTF) \citep{acar2011all} was  proposed for multimodal data fusion. CMTF estimates the underlying latent factors for both tensor and matrix data simultaneously by taking the coupling between tensor and matrix data into account. This feature makes CMTF a promising model in analyzing heterogeneous data, which generally have different structures and modalities. 
 
During latent factor estimation, CMTF solves an objective function that approximates a CP decomposition for the tensor modality and a singular value decomposition for the second modality with the assumption that the factors from one mode of each modality are the same. Given $\ts{X}_{1} \in \mathbb{R}^{I_{1} \times I_2 \times \ldots \times I_{d}}$ and $\pmb{X}_{2} \in \mathbb{R}^{I_{1} \times J_2}$, without loss of generality assume that the factors from the first mode of the tensor $\ts{X}_1$ span the column space of the matrix $\pmb{X}_2$. CMTF then tries to estimate all factors by minimizing:
\begin{equation}
    \pmb{Q}(\kruskal{U}_{1}, \pmb{V}) = \frac{1}{2} \|\ts{X}_{1} - \llbracket \pmb{X}_{1}^{(1)}, \pmb{X}_1^{(2)}, ... \pmb{X}_{1}^{(d)} \rrbracket\|^{2}_{\text{Fro}} + \frac{1}{2} \| \pmb{X}_{2} - \pmb{X}_2^{(1)} \pmb{X}_2^{(2)T}\|^{2}_{\text{Fro}}, \quad \text{s.t.} \;\pmb{X}_1^{(1)}=\pmb{X}_2^{(1)},
     \label{equ:CMTF_}
\end{equation}
where $\pmb{X}_{p}^{(m)}$ are the factor matrices for modality $p$ and mode $m$. The factor matrices $\pmb{X}_1^{(1)}=\pmb{X}_2^{(1)}$ are the coupled factors between tensor and matrix data. An illustration of this coupling is given in Figure \ref{fig:data_couple_}. 
These factor matrices can also be represented in Kruskal form, $\kruskal{U}_1=\llbracket \pmb{X}_{1}^{(1)}, \pmb{X}_1^{(2)}, ... \pmb{X}_{1}^{(d)} \rrbracket$ and $\kruskal{U}_2=\llbracket\pmb{X}_2^{(1)}, \pmb{X}_2^{(2)}\rrbracket$. By minimizing the objective function $\pmb{Q}(\kruskal{U}_{1}, \kruskal{U}_2)$, CMTF estimates latent factors for the tensor and matrix data jointly which allows it to utilize information from both modalities. \citet{acar2011all} uses a gradient descent algorithm to optimize the objective function (\ref{equ:CMTF_}). Although this model is formulated for the joint decomposition of a $d$th order tensor and a matrix, extensions to two or more tensors with  couplings across multiple modes are possible. 

In real data, couplings across different modalities might include shared or modality-specific (individual) components. Shared components correspond to those columns of the factor matrices that contribute to the decomposition of both modalities, while individual components carry information unique to the corresponding modality. Although CMTF provides a successful framework for joint data analysis, it often fails to obtain a unique estimation for shared or individual components. As a result, any further statistical analysis and learning from CMTF estimation will suffer from the uncertainty in latent factors. To address this issue, \citet{acar2017acmtf} proposed Advanced Coupled Matrix Tensor Factorization (ACMTF) by introducing a sparsity penalty to the weights of latent factors in the objective function (\ref{equ:CMTF_}), and restricting the norm of the columns of the factors to be unity to provide uniqueness up to a permutation. This modification provides a more precise estimation for latent factors compared to CMTF (\citet{acar2017acmtf, acar2017tensor}). In our framework, we utilize ACMTF to extract the latent factors which are in turn used to build a classifier for multimodal data.

\subsection{CP-STM for Tensor Classification}
CP-STM has been previously studied by \citet{tao2005supervised, he2014dusk, he2017kernelized} and uses CP tensor to construct STM types of model. Assume there is a collection of data $T_{n} = \{ (\ts{X}_{1}, y_{1}), (\ts{X}_{2}, y_{2}), ..., (\ts{X}_{n}, y_{n})\}$, where $\ts{X}_{i} \in \nspace{X} \subset \mathbb{R}^{{I}_{1} \times I_{2} \times ... \times I_{d}}$ are $d$-way tensors. $\nspace{X}$ is a compact tensor space, which is a subspace of  $\mathbb{R}^{{I}_{1} \times I_{2} \times ... \times I_{d}}$. $y_{i} \in \{1, -1 \} $ are binary labels. CP-STM assumes the tensor predictors are in CP format, and can be classified by the function which minimizes the objective function 
\begin{equation}
    \min \quad \lambda||\pmb{f}||^{2} + \frac{1}{n} \sum \limits_{i = 1}^{n}\mathcal{L}(\pmb{f}(\ts{X}_{i}), y_{i}).
    \label{equ:CP-STM}
\end{equation}
By using tensor kernel function 
\begin{equation}
    K(\ts{X}_{1}, \ts{X}_{2}) = \sum \limits_{l, m = 1} ^{r} \prod_{j = 1}^{d} K^{(j)}(\pmb{x}_{1, l}^{(j)}, \pmb{x}_{2, m}^{(j)}),
    \label{equ:TK}
\end{equation}
where $\ts{X}_{1} = \sum \limits_{l = 1}^{r}\pmb{x}_{1l}^{(1)} \circ .. \circ \pmb{x}_{1l}^{(d)}$ and $\ts{X}_{2} = \sum \limits_{l = 1}^{r}\pmb{x}_{2l}^{(1)} \circ .. \circ \pmb{x}_{2l}^{(d)}$. The STM classifier can be written as 
\begin{equation}
    \pmb{f} (\ts{X}) = \sum \limits_{i = 1}^{n}\alpha_{i} y_{i}K(\ts{X}_{i}, \ts{X}) = \pmb{\alpha}^{T}\pmb{D}_{y}\pmb{K}(\ts{X})
    \label{equ:STM}
\end{equation}
where $\ts{X}$ is a new $d$-way rank-$r$ tensor, $\pmb{\alpha} = [\alpha_{1}, ..., \alpha_{n}]^{T}$ is the coefficient vector, $\pmb{D}_{y}$ is a diagonal matrix whose diagonal elements are $y_{1}, .., y_{n}$ and $\pmb{K}(\ts{X}) = [K(\ts{X}_{1}, \ts{X}), ..., K(\ts{X}_{n}, \ts{X})]^{T}$ is a column vector, whose values are kernel values computed between training and   test data. We denote the collection of functions in the form of (\ref{equ:STM}) with $\nspace{H}$, which is a functional space also known as Reproducing Kernel Hilbert Space (RKHS). The optimal classifier CP-STM $\pmb{f} \in \nspace{H}$ can be estimated by plugging function (\ref{equ:STM}) into objective function (\ref{equ:CP-STM}), and minimize it with Hinge or Squared Hinge loss. The coefficients of the optimal CP-STM model is denoted by $\pmb{\alpha}^{*}$. The classification model is statistically consistent if the tensor kernel function satisfying the universal approximating property, which is shown by \citet{li2019universal}.

\subsection{Multiple Kernel Learning}
Multiple kernel learning (MKL) creates new kernels using a linear or non-linear combination of single kernels to measure inner products between data. Statistical learning algorithms such as support vector machine and kernel regression can then utilize the new combined kernels instead of single kernels to obtain better learning results and avoid the potential bias from kernel selection (\citet{gonen2011multiple}). A more important and  related reason for using MKL is that different kernels can take inputs from various data representations possibly from different sources or modalities. Thus, combining kernels and using MKL is one possible way of integrating multiple information sources. 

Given a collection of kernel functions $\{ K_{1}(\cdot, \cdot), ... K_{m}(\cdot, \cdot) \}$, a new kernel function can be constructed by 
\begin{equation}
    K(\cdot, \cdot) = \pmb{f_{\eta}} (\{ K_{1}(\cdot, \cdot), ... K_{m}(\cdot, \cdot) \} | \pmb{\eta})
    \label{equ:MKL_}
\end{equation}
where $\pmb{f_{\eta}}$ is a linear or non-linear function and $\pmb{\eta}$ is a vector whose elements are the weights for the kernel combination. Linear combination methods are the most popular in multiple kernel learning, where the kernel function is parameterized as 
\begin{equation}
    \begin{split}
        K(\cdot, \cdot) & = \pmb{f_{\eta}} (\{ K_{1}(\cdot, \cdot), ... K_{m}(\cdot, \cdot) \} | \pmb{\eta})\\
        & = \sum \limits_{l = 1}^{m} \eta_{l}K_{l}(\cdot, \cdot). 
    \end{split}
    \label{equ:MKL_linear_}
\end{equation}
The weight parameters $\eta_{l}$ can be simply assumed to be the same (unweighted) (\citet{pavlidis2001gene, ben2005kernel}), or be determined by looking at some performance measures for each kernel or data representation (\citet{tanabe2008simple, qiu2008framework}).  There are few more advanced approaches such as optimization-based, Bayesian approaches, and boosting approaches that can also be adopted (\citet{lanckriet2004learning, fung2004fast, varma2007learning, kloft2009efficient, girolami2007data, christoudias2009bayesian, bennett2002mark}). In this work, we only consider linear combination (\ref{equ:MKL_linear_}), and select the weight parameters in a heuristic data-driven way to construct our C-STM model.

\section{Methodology}
\label{sec:methodology_}
\begin{figure}
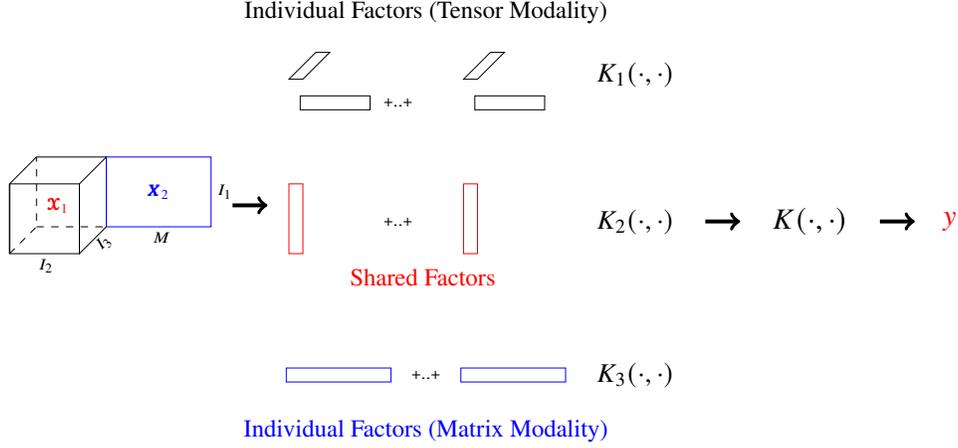

    \centering
    \resizebox{0.9\textwidth}{!}{
    \cstmmodelpipeline} 
    \caption{C-STM Model Pipeline}
    \label{fig:modelpipeline_}
\end{figure}

Let $T_{n} = \{ (\ts{X}_{1, 1}, \pmb{X}_{1, 2}, y_{1}), ..., (\ts{X}_{n, 1}, \pmb{X}_{n, 2}, y_{n}) \}$ be training data, where each sample $t\in \{1,\dots,n\}$ has two data modalities $\pmb{X}_{t,1}, \pmb{X}_{t,2}$ and a corresponding binary label $y_{t}\in \{1, -1 \}$. In this work, following \citet{acar2011all}, we assume that the first data modality is a third-order tensor, $\ts{X}_{t,1}\in\mathbb{R}^{I_1 \times I_2 \times I_3}$, and the other is a matrix, $\pmb{X}_{t,2}\in\mathbb{R}^{I_{4}\times I_{3}}$. The third mode of $\ts{X}_{t,1}$ and the second mode of $\ts{X}_{t,2}$ are assumed to be coupled for each $t$, i.e. the factor matrix is assumed to be fully or partially shared across these modes. Utilizing this coupling, one can extract factors that better represent the underlying structure of the data, and preserve and utilize the discriminative power of the factors from both modalities. Our approach, C-STM, consists of two stages: Multimodal tensor factorization, i.e., ACMTF, and coupled support tensor machine as illustrated in Figure \ref{fig:modelpipeline_}. In this section, we present both stages and the corresponding procedures.

\subsection{Multimodal Tensor Factorization}
In this work, the first aim is to perform a joint factorization across two modalities for each training sample, $t$. Let $\kruskal{U}_{t,1} = \llbracket \pmb{\zeta}; \pmb{X}_{t,1}^{(1)},\pmb{X}_{t,1}^{(2)},\pmb{X}_{t,1}^{(3)} \rrbracket$ denote the Kruskal tensor of $\ts{X}_{t,1}$, and $\kruskal{U}_{t, 2} = \llbracket \pmb{\sigma}; \pmb{X}_{t, 2}^{(1)}, \pmb{X}_{t, 2}^{(2)} \rrbracket$ denote the singular value decomposition of $\pmb{X}_{t,2}$. The weights of the columns of each factor matrix $\pmb{X}_{t,p}^{(m)}$, where $p$ is the index for modality and $m$ denotes the mode, are denoted by $\pmb{\zeta}$ and $\pmb{\sigma}$ and the norms of these columns are constrained to be 1 to avoid redundancy. The objective function of ACMTF \citep{acar2014structure, acar2017acmtf} is then given by:
\begin{equation}
    \begin{split}
        \pmb{Q}(\kruskal{U}_{t, 1}, \kruskal{U}_{t, 2}) & = \gamma \|\ts{X}_{t,1}- \llbracket \pmb{\zeta}; \pmb{X}_{t,1}^{(1)},\pmb{X}_{t,1}^{(2)},\pmb{X}_{t,1}^{(3)} \rrbracket \|_{\text{Fro}}^2+ \gamma\|\pmb{X}_{t,2} - \pmb{X}_{t,2}^{(1)} \pmb{\Sigma} \pmb{X}_{t,2}^{(2)^\top}\|_{\text{Fro}}^2\\ &
         + \beta\|\pmb{\zeta}\|_0+\beta\|\pmb{\sigma}\|_0 \\
        \text{s.t.} \quad & \pmb{X}_{t,1}^{(3)} = \pmb{X}_{t,2}^{(2)}\\
        & \|\pmb{x}^{(1)}_{t,1,k}\|_2=\|\pmb{x}^{(2)}_{t,1,k}\|_2=\|\pmb{x}^{(3)}_{t,1,k}\|_2=    \|\pmb{x}^{(1)}_{t,2,k}\|_2= \|\pmb{x}^{(2)}_{t,2,k}\|_2=1, \\
        & \quad \forall k\in \{1,\dots,r\}
    \end{split}
    \label{equ:ACMTF_}
\end{equation}
where $\pmb{\Sigma}$ is a diagonal matrix whose elements are the singular values $\pmb{\sigma}$ of the matrix $\pmb{X}_{t, 2}$ and $\pmb{x}_{t, m, k}^{(j)} \in \mathbb{R}^{I_{j}}$ denotes the columns of the factor matrices for the object $\ts{X}_{t, m}$. The objective function in \eqref{equ:ACMTF_} includes penalties for the number of non-zero weights in both tensor and matrix decomposition. Thus, the model identifies the shared and individual components. These factors are then considered as different data representations for multimodal data, and used to predict the labels $y_{t}$ in C-STM classifier. 

\subsection{Coupled Support Tensor Machine (C-STM)}
C-STM uses the idea of multiple kernel learning and considers the coupled and uncoupled factors from ACMTF decomposition as various data representations. As a result, we use three different kernel functions to measure their similarity, i.e., inner products. One can think of these three kernels inducing three different feature maps transforming multimodal factors into different feature spaces. In each feature space, the corresponding kernel measures the similarity between factors in this specific data modality. The similarities of multimodal factors are then integrated by combining the kernel measures through a linear combination. This procedure is illustrated in  Figure \ref{fig:modelpipeline_}. In particular,  kernel $K_{1}$ is a tensor kernel (equation (\ref{equ:TK})) since the first individual factors are tensor CP factors. The kernel function for C-STM is defined as 
\begin{gather}
       K\bigg( (\ts{X}_{t, 1}, \pmb{X}_{t, 2}) ,  (\ts{X}_{i, 1}, \pmb{X}_{i, 2})\bigg)  = K\bigg( (\kruskal{U}_{t, 1}, \kruskal{U}_{t, 2}), (\kruskal{U}_{i, 1}, \kruskal{U}_{i, 2}) \bigg) \nonumber \\
        =  \sum \limits_{k, l = 1}^{r} w_{1}K_{1}^{(1)} (\pmb{x}_{t, 1, k}^{(1)}, \pmb{x}_{t, 1, l}^{(1)}) K_{1}^{(2)} (\pmb{x}_{t, 1, k}^{(2)}, \pmb{x}_{t, 1, l}^{(2)})
        + w_{2}K_{2} (\pmb{x}_{t, 1, k}^{(3)*}, \pmb{x}_{t, 1, l}^{(3)*}) + w_{3}K_{3} (\pmb{x}_{t, 2, k}^{(1)}, \pmb{x}_{t, 2, l}^{(1)})
    \label{equ:C_TK_}
\end{gather}
for two pairs of decomposed tensor matrix factors $(\kruskal{U}_{t, 1}, \kruskal{U}_{t, 2})$ and $(\kruskal{U}_{i, 1}, \kruskal{U}_{i, 2})$. $\pmb{x}_{t, 1, k}^{(3)*}$ is the average of the estimated shared factors $ \frac{1}{2}[\pmb{x}_{t, 1, k}^{(3)} + \pmb{x}_{t, 2, k}^{(2)}]$. $w_{1}$, $w_{2}$, and $w_{3}$ are the three weights parameters combining three kernel functions, which can be tuned by cross-validation. 

With the kernel function in (\ref{equ:C_TK_}), C-STM model tries to estimate a bivariate decision function $\pmb{f}$ from a collection of functions $\nspace{H}$ such that 
\begin{equation}
      \pmb{f} = \arg \min \quad \lambda \cdot ||\pmb{f}||^{2} + \frac{1}{n} \sum \limits_{i = 1}^{n}\mathcal{L}(\pmb{f}(\ts{X}_{i}), y_{i}),
      \label{equ:C-STM_obj_}
\end{equation}
where $\mathcal{L}(\ts{X}_{i}, y_{i}) = \max \big( 0, 1 - \pmb{f}(\ts{X}_{i}) \cdot y_{i} \big)$ is Hinge loss. $\nspace{H}$ is defined as the collection of all functions in the form of 
\begin{equation}
   \begin{split}
       \pmb{f} (\ts{X}_{1}, \pmb{X}_{2}) & = \sum \limits _{t = 1}^{n}\alpha_{i}y_{i} K((\ts{X}_{t, 1}, \pmb{X}_{t, 2}), (\ts{X}_{1}, \pmb{X}_{2}))\\
        & = \pmb{\alpha}^{T}\pmb{D}_{y}\pmb{K}(\ts{X}_{1}, \pmb{X}_{2})
   \end{split}
    \label{equ:C_STM_decision_}
\end{equation}
due to the well-known representer theorem (\citet{argyriou2009there}) for any pair of testing data $(\ts{X}_{1}, \pmb{X}_{2})$ and for $\pmb{\alpha} \in \mathbb{R}^{n}$. For all possible values of $\pmb{\alpha}$, equation (\ref{equ:C_STM_decision_}) defines the data collection $\nspace{H}$. $\pmb{D}_{y}$ is a diagonal matrix whose diagonal elements are labels from the training data $T_{n}$. 
$\pmb{K}(\ts{X}_{1}, \pmb{X}_{2})$ is a $n \times 1$  vector whose $t$-th element is $ K\big( (\ts{X}_{t, 1}, \pmb{X}_{t, 2}) ,  (\ts{X}_{1}, \pmb{X}_{2})\big)$. The optimal C-STM decision function, denoted by $\pmb{f_{n}} =  \pmb{\alpha}^{*T}\pmb{D}_{y}\pmb{K}(\ts{X}_{1}, \pmb{X}_{2})$, can be estimated by solving the quadratic programming problem 
\begin{equation}
    \begin{split}
       & \underset{\pmb{\alpha} \in \mathbb{R}^{n}}{\min} \quad \frac{1}{2}\pmb{\alpha}^{T}\pmb{D}_{y}\pmb{K}\pmb{D}_{y}\pmb{\alpha} - \pmb{1}^{T}\pmb{\alpha},\\
       & \text{S.T.} \quad \pmb{\alpha}^{T}\pmb{y} = 0,\\
       & \quad \quad \quad 0 \preceq \pmb{\alpha} \preceq \frac{1}{2n \lambda},
    \end{split}
    \label{equ:QP_}
\end{equation}
where $\pmb{K}$ is the kernel matrix constructed by function (\ref{equ:C_TK_}). Problem (\ref{equ:QP_}) is the dual problem of (\ref{equ:C-STM_obj_}), and its optimal solution $\pmb{\alpha}^{*}$ also minimizes the objective function (\ref{equ:C-STM_obj_}) when plugging functions in the form of (\ref{equ:C_STM_decision_}). For a new pair of test points $(\ts{X}_{1}, \pmb{X}_{2})$, the class label is predicted as $\text{Sign} \big[ \pmb{f_{n}} (\ts{X}_{1}, \pmb{X}_{2})   \big]$.

\section{Model Estimation}
\label{sec:estimation_}
In this section, we first present the estimation procedure for coupled tensor matrix decomposition (\ref{equ:ACMTF_}), and then combine it with the classification procedure to summarize the algorithm for C-STM.

To satisfy the constraints in the objective function (\ref{equ:ACMTF_}), the function $\pmb{Q}(\kruskal{U}_{t, 1}, \kruskal{U}_{t, 2})$ is converted to a differentiable and unconstrained form given by: 
\begin{equation}
    \begin{split}
        \pmb{Q}(\kruskal{U}_{t, 1}, \kruskal{U}_{t, 2}) = & \gamma\|\ts{X}_{t,1} - \llbracket \pmb{\zeta}; \pmb{X}_{t,1}^{(1)},\pmb{X}_{t,1}^{(2)}, \pmb{X}_{t,1}^{(3)} \rrbracket\|_{\text{Fro}}^2 + \gamma\|\ts{X}_{t,2} - \pmb{X}_{t,2}^{(1)} \pmb{\Sigma} \pmb{X}_{t,2}^{(2)^\top}\|_{\text{Fro}}^2 \\
        & + \xi \|\pmb{X}_{t,1}^{(3)} - \pmb{X}_{t,2}^{(2)}\|_{\text{Fro}}^2\\ 
        & + \sum_{k=1}^{r}  \bigg[ \beta\sqrt{\zeta_r^2+\epsilon} + \beta\sqrt{\sigma_r^2+\epsilon}  + \theta \big[ (\|\ts{X}^{(1)}_{t,1,k}\|_2-1)^2+(\|\ts{X}^{(2)}_{t,1,k}\|_2-1)^2 \\
        & + (\|\ts{X}^{(3)}_{t,1,k}\|_2-1)^2 
         + (\|\ts{X}^{(1)}_{t,2,k}\|_2-1)^2 + (\|\ts{X}^{(2)}_{t,2,k}\|_2-1)^2 \big] \bigg],
    \end{split}
    \label{equ:unconstrained_ACMTF_}
\end{equation}
where $\ell_{1}$ norm penalties in (\ref{equ:ACMTF_}) are replaced with differentiable approximations; $\xi$ and $\theta$ are Lagrange multipliers and $\epsilon > 0$ is a very small number. This unconstrained optimization problem can be solved by nonlinear conjugate gradient descent (\citet{acar2011all, acar2014structure, mosayebi2020correlated}). 

Let $\ts{T}_{t}$ be the full (created by converting Kruskal tensor, or the factor matrices into multidimensional array form) tensor of $\kruskal{U}_{t, 1}$, and $\pmb{M}_{t} = \pmb{X}_{t,2}^{(1)} \pmb{\Sigma} \pmb{X}_{t,2}^{(2)^\top}$, the partial derivative of each latent factor can be derived as follows:
\begin{equation}
     \frac{\delta \pmb{Q}(\kruskal{U}_{t, 1}, \kruskal{U}_{t, 2})}{\delta \pmb{X}_{t,1}^{(1)}} = \gamma(\ts{T}_{t} - \ts{X}_{t,1})_{(1)} (\pmb{\zeta}^\top\odot \pmb{X}_{t,1}^{(3)} \odot \pmb{X}_{t,1}^{(2)}) + \theta (\pmb{X}_{t,1}^{(1)} - \bar{\pmb{X}}_{t,1}^{(1)}),
\end{equation}
\begin{equation}
    \gamma\frac{\delta \pmb{Q}(\kruskal{U}_{t, 1}, \kruskal{U}_{t, 2}) }{\delta \pmb{X}_{t,1}^{(2)}} = (\ts{T}_{t} - \ts{X}_{t,1})_{(2)} (\pmb{\zeta}^\top\odot X_{t,1}^{(3)}\odot \pmb{X}_{t,1}^{(1)}) + \theta(\pmb{X}_{t,1}^{(2)} - \bar{\pmb{X}}_{t,1}^{(2)}),
\end{equation}
\begin{equation}
    \gamma\frac{\delta \pmb{Q}(\kruskal{U}_{t, 1}, \kruskal{U}_{t, 2}) }{\delta \pmb{X}_{t,1}^{(3)}} = (\ts{T}_{t} - \ts{X}_{t,1})_{(3)} (\pmb{\zeta}^\top\odot X_{t,1}^{(2)}\odot \pmb{X}_{t,1}^{(1)}) + \xi (\pmb{X}_{t, 1}^{(3)} - \pmb{X}_{t, 2}^{(2)})+  \theta(\pmb{X}_{t,1}^{(3)} - \bar{\pmb{X}}_{t,1}^{(3)}),
\end{equation}
\begin{equation}
    \gamma\frac{\delta \pmb{Q}(\kruskal{U}_{t, 1}, \kruskal{U}_{t, 2}) }{\delta \pmb{X}_{t,2}^{(1)}} = (\pmb{M}_{t} - \pmb{X}_{t,2}) \pmb{X}_{t, 2}^{(2)} \pmb{\Sigma} +  \theta(\pmb{X}_{t,2}^{(1)} - \bar{\pmb{X}}_{t,2}^{(1)}),
\end{equation}
\begin{equation}
    \gamma\frac{\delta \pmb{Q}(\kruskal{U}_{t, 1}, \kruskal{U}_{t, 2}) }{\delta \pmb{X}_{t,2}^{(2)}} = (\pmb{M}_{t} - \pmb{X}_{t,2})^{\top} \pmb{X}_{t, 2}^{(1)} \pmb{\Sigma} + \tau ( \pmb{X}_{t, 2}^{(2)} - \pmb{X}_{t, 1}^{(3)}) +  \theta(\pmb{X}_{t,2}^{(2)} - \bar{\pmb{X}}_{t,2}^{(2)}),
\end{equation}
\begin{equation}
    \frac{\delta \pmb{Q}(\kruskal{U}_{t, 1}, \kruskal{U}_{t, 2}) }{\delta \sigma_{k}} =  \pmb{x}_{t, 2, k}^{(1) \top} (\pmb{M}_{t} - \pmb{X}_{t, 2}) \pmb{x}_{t, 2, k}^{(2) } + \frac{\beta}{2} \frac{\sigma_{k}}{\sqrt{\sigma_{k}^{2} + \epsilon}}, \; k \in \{1,..., r\},
\end{equation}
\begin{equation}
    \frac{\delta \pmb{Q}(\kruskal{U}_{t, 1}, \kruskal{U}_{t, 2}) }{\delta \zeta_{k}} = \text{vec}(\ts{T}_{t} - \ts{X}_{t,1})^\top \left(\pmb{x}_{t, 1, k}^{(3)}\odot \pmb{x}_{t, 1, k}^{(2)} \odot \pmb{x}_{t, 1, k}^{(1)}\right) + \frac{\beta}{2} \frac{\zeta_{k}}{\sqrt{\zeta_{k}^{2} + \epsilon}}, \; k \in \{1,..., r\},
\end{equation}
where $\text(vec(.))$ is a vectorization operator that stacks all elements of the operand in a column vector, $\ts{T}_{(j)}$ denotes the mode-$j$ unfolding of a tensor $\ts{T}$, and $\odot$ denotes Khatri-Rao product.  $\bar{\pmb{M}}$ is a normalized matrix whose columns have unit $\ell_{2}$ norms. 

By combining all of the partial derivatives, the partial derivative of the objective function is given by: 
\begin{equation}
\begin{split}
    \pmb{\triangledown Q}(\kruskal{U}_{t, 1}, \kruskal{U}_{t, 2}) = \bigg[ & \frac{\delta \pmb{Q}(\kruskal{U}_{t, 1}, \kruskal{U}_{t, 2})}{\delta \pmb{X}_{t,1}^{(1)}}, 
        \frac{\delta \pmb{Q}(\kruskal{U}_{t, 1}, \kruskal{U}_{t, 2}) }{\delta \pmb{X}_{t,1}^{(2)}},
         \frac{\delta \pmb{Q}(\kruskal{U}_{t, 1}, \kruskal{U}_{t, 2}) }{\delta \pmb{X}_{t,1}^{(3)}},\\
        & \frac{\delta \pmb{Q}(\kruskal{U}_{t, 1}, \kruskal{U}_{t, 2}) }{\delta \pmb{X}_{t,2}^{(2)}},
        \frac{\delta \pmb{Q}(\kruskal{U}_{t, 1}, \kruskal{U}_{t, 2}) }{\delta \zeta_{1}},
        ...
         \frac{\delta \pmb{Q}(\kruskal{U}_{t, 1}, \kruskal{U}_{t, 2}) }{\delta \sigma_{1}},
        ...\bigg]^{\top}
\end{split}
    \label{equ_ACMTF_partialder_}
\end{equation}
which is a $2r+5$ dimensional vector. As mentioned in \cite{acar2014structure}, a nonlinear conjugate gradient method with Hestenes-Stiefel updates is used to optimize (\ref{equ:unconstrained_ACMTF_}). The procedure is described in Algorithm \ref{alg:ACMTF_}. 

\begin{algorithm}[t]
\caption{ACMTF Decomposition}
\label{alg:ACMTF_}
\begin{algorithmic}[1]
\Procedure{ACMTF} {} 
\State \textbf{Input:} Multimodal data $(\ts{X}_{1}, \pmb{X}_{2})$, r,  $\eta$, S (Upper limit for the number of iterations)
\State \textbf{Output:} $\kruskal{U}_{t, 1}^{*}, \kruskal{U}_{t, 2}^{*}$ 
    \State $\kruskal{U}_{t, 1}, \kruskal{U}_{t, 2} = \kruskal{U}_{t, 1}^{0}, \kruskal{U}_{t, 2}^{0}$ \Comment{Initial value}
    \State $\pmb{\Delta}_{0} = -\pmb{\triangledown Q}(\kruskal{U}_{t, 1}^{0}, \kruskal{U}_{t, 2}^{0})$
    \State $\varphi_{0} = \arg \min_{\varphi} \pmb{Q} \big[ (\kruskal{U}_{t, 1}^{0}, \kruskal{U}_{t, 2}^{0}) + \varphi \pmb{\Delta}_{0}\big]$
    \State $\kruskal{U}_{t, 1}^{1}, \kruskal{U}_{t, 2}^{1} = (\kruskal{U}_{t, 1}^{0}, \kruskal{U}_{t, 2}^{0}) + \varphi_{0}\pmb{\Delta}_{0}$
    \State $\pmb{g}_{0} = \pmb{\Delta}_{0}$
    \While{s $<$ S and $\|\pmb{Q} (\kruskal{U}_{t, 1}^{s}, \kruskal{U}_{t, 2}^{s}) - \pmb{Q} (\kruskal{U}_{t, 1}^{s - 1}, \kruskal{U}_{t, 2}^{s - 1})\| \geqslant \eta$}
     \State $\pmb{\Delta}_{s + 1} = -\pmb{\triangledown Q}(\kruskal{U}_{t, 1}^{s}, \kruskal{U}_{t, 2}^{s})$
     \State $\pmb{g}_{s + 1} = \pmb{\Delta}_{s + 1} + \frac{\pmb{\Delta}_{s + 1}^{\top}(\pmb{\Delta}_{s + 1} - \pmb{\Delta}_{s})}{-\pmb{g}_{s}^{\top}(\pmb{\Delta}_{s + 1} - \pmb{\Delta}_{s})}\pmb{g}_{s}$
    \State $\varphi_{s + 1} = \arg \min_{\varphi} \pmb{Q} \big[ (\kruskal{U}_{t, 1}^{s}, \kruskal{U}_{t, 2}^{s}) + \varphi \pmb{g}_{s + 1}\big]$
    \State $\kruskal{U}_{t, 1}^{s+1}, \kruskal{U}_{t, 2}^{s+1} = (\kruskal{U}_{t, 1}^{s}, \kruskal{U}_{t, 2}^{s}) + \varphi_{s+1}\pmb{g}_{s + 1}$
    \EndWhile
\EndProcedure
\end{algorithmic}
\end{algorithm}

Once the factors for all data pairs in the training set $T_{n}$ are extracted, we can create the kernel matrix using the kernel function in \eqref{equ:C_TK_}. By solving the quadratic programming problem (\ref{equ:QP_}), we can obtain the optimal decision function $\pmb{f_{n}}$. This two-stage procedure for C-STM estimation is summarized in  Algorithm \ref{alg:C-STM_}.
\begin{algorithm}[t]
\caption{Coupled Support Tensor Machine}
\label{alg:C-STM_}
\begin{algorithmic}[1]
\Procedure{C-STM} {} 
\State \textbf{Input:} Training set $T_{n} = \{ (\ts{X}_{1, 1}, \pmb{X}_{1, 2}, y_{1}), ..., (\ts{X}_{n, 1}, \pmb{X}_{n, 2}, y_{n}) \}$, $\pmb{y}$, kernel function $K$, $r$, $\lambda$, $\eta$, S
\For{t = 1, 2,...n}
    \State $\kruskal{U}_{t, 1}^{*}, \kruskal{U}_{t, 2}^{*}$ = ACMTF$\left((\ts{X}_{t, 1}, \pmb{X}_{t, 2}), r, \eta, \text{S}\right)$
\EndFor

\State Create initial matrix $\pmb{K} \in \mathbb{R}^{n \times n}$
\For{t = 1,...,n}
    \For{i = 1,...,i}
        \State $\pmb{K}[i, t] = K\bigg( (\kruskal{U}_{t, 1}, \kruskal{U}_{t, 2}), (\kruskal{U}_{i, 1}, \kruskal{U}_{i, 2}) \bigg) $ \Comment{Kernel values}
            \State $\pmb{K}[i, t] = \pmb{K}[t, i]$
    \EndFor
\EndFor
\State Solve the quadratic programming problem (\ref{equ:QP_}) and find the optimal $\pmb{\alpha}^{*}$.
\State \textbf{Output:} $\pmb{\alpha}^{*}$ 
\EndProcedure
\end{algorithmic}
\end{algorithm}

\section{Theory}
\label{sec:theory_}
In this section, we provide some preliminary theoretical results to validate the  C-STM model. The first proposition provides a sketch of proof of convergence for the coupled matrix tensor decomposition. 
\begin{proposition}
Suppose for every pair of multimodal data $\ts{X}_{t, 1}$ and $\pmb{X}_{t, 2}$, the optimal latent factor estimate is the optimal solution for the objective function (\ref{equ:ACMTF_}), which is denoted by $\kruskal{U}_{t, 1}^{*}, \kruskal{U}_{t, 2}^{*}$. The conjugate gradient descent algorithm \ref{alg:ACMTF_} converges to stable estimates for tensor and matrix components $\kruskal{U}_{t, 1}^{*}, \kruskal{U}_{t, 2}^{*}$ where:
\begin{gather}
    D\left(\left(\kruskal{U}_{t,1}^{\tau},\kruskal{U}_{t,2}^{\tau}\right), \left(\kruskal{U}_{t,1}^{*},\kruskal{U}_{t,2}^{*}\right)\right)\xrightarrow{}0,\quad \text{as}\; \tau\xrightarrow{}\infty,
\end{gather}
where $\tau$ is the number of iterations and $D(.,.)$ is a distance measure between Kruskal tensor sets such as $\ell_2$ distance between factors with an appropriate selection of permutations.
\label{prop:ACMTF_converge_}
\end{proposition}
\noindent This proposition is a direct result of the convergence property of nonlinear conjugate gradient descent algorithm with line search (\citet{nocedal2006numerical,hager2006survey, zoutendijk1970nonlinear, wolfe1969convergence, wolfe1971convergence}). The convergence rate of non-linear conjugate gradient descent is linear. For detailed convergence properties of nonlinear conjugate gradient descent with Hestenes-Stiefel updates on non-convex objectives, the readers are referred to \citet{nocedal2006numerical, powell1984nonconvex}. 

The next result discusses the statistical property of C-STM. Let's assume the risk of a decision function, $\pmb{f}$, is $\mathcal{R}(\pmb{f}) = \mathbb{E}_{\nspace{X} \times \nspace{Y}} \big[ \pmb{1}\{\pmb{f}(\ts{X}) \neq y \} \big]$, where $\nspace{X} \subset \mathbb{R}^{I_{1} \times .. \times I_{d}}$ is a subspace of $\mathbb{R}^{I_{1} \times .. \times I_{d}}$. $\nspace{Y} = \{1, -1 \}$. The expectation is taken over the joint distribution defined on $\nspace{X} \times \nspace{Y}$, which is a data domain. The function $\pmb{1} \{ \cdot \}$ is an indicator function measuring the loss of classification function $\pmb{f}$. It is also known as the "zero-one" loss since its value is zero when the decision function provides correct prediction and is one otherwise. If there is a $\pmb{f}^{*} : \nspace{X} \rightarrow \nspace{Y}$ from the collection of all measurable functions such that $\pmb{f}^{*} = \arg \min \mathcal{R}(\pmb{f})$, its risk is called the Bayes risk for the classification problem with data from $\nspace{X} \times \nspace{Y}$. We denote the Bayes risk as $\mathcal{R}^{*} = \mathcal{R}(\pmb{f}^{*})$. With different training sets $T_{n}$, we can estimate a sequence of decision functions $\pmb{f_{n}}$ under the same training procedure. This sequence of decision function $\{ \pmb{f_{n}}\}$ is called a decision rule. Obviously, C-STM is a decision rule if different training sets are provided. A decision rule is statistically consistent if $\mathcal{R}(\pmb{f_{n}})$ converges to the Bayes risk $\mathcal{R}^{*}$ as the size of training data $n$ increases (see eg. \citet{devroye2013probabilistic}). The consistency property is desirable for classification rules, because a consistent rule guarantees to reconstruct the whole data distribution with more training data / observations. The reconstruction here means the Bayes risk of the classification problem will be eventually the same as the risk of estimated classifier with sufficient training data, and thus will be known. Our next result shows that C-STM is a statistically consistent decision rule. 

\begin{proposition}
Given the tensor and matrix factors for all data in the domain, the classification risk of C-STM, $\mathcal{R}(\pmb{f_{n}})$, converges to the optimal Bayes risk almost surely, i.e.
$$\mathcal{R}(\pmb{f_{n}}) \rightarrow \mathcal{R}^{*} \quad a.s.   \quad n \rightarrow \infty$$
if the following conditions are satisfied:
\begin{enumerate}[label=\textbf{AS.\arabic*}]
    \item \label{cond:A1_} The loss function $\mathcal{L}$ is self-calibrated (see \cite{steinwart2008support}), and is $C(W)$ local Lipschitz continuous in the sense that for $|a| \leqslant W < \infty$ and $|b|\leqslant W < \infty$
    \begin{equation*}
        |\mathcal{L}(a, y) - \mathcal{L}(b, y)| \leqslant C(W) |a - b| 
    \end{equation*}
    In addition, we need $\underset{y \in \{1, -1\}}{\sup}\mathcal{L}(0, y) \leqslant L_{0} < \infty$.

    \item \label{cond:A2_} The kernel functions $K_{1}^{(1)}(\cdot, \cdot)$, $K_{1}^{(2)}(\cdot, \cdot)$,  $K_{2}(\cdot, \cdot)$, and $K_{3}(\cdot, \cdot)$ used to compose the coupled tensor kernel (\ref{equ:C_TK_}) are regular vector-based kernels satisfying the universal approximating property. A kernel has this property if it  satisfies the following condition.  Suppose $\nspace{X}$ is a compact subset of the Euclidean space $\mathbb{R}^{p}$, and $C(\nspace{X}) = \{ \pmb{f}: \nspace{X} \rightarrow \mathbb{R} \} $ is the collection of all continuous functions defined on $\nspace{X}$. The kernel function is also defined on $\nspace{X} \times \nspace{X}$, and its reproduction kernel Hilbert space (RKHS) is  $\nspace{H}$. Then $\forall \pmb{g} \in C(\nspace{X})$, $\exists \pmb{f} \in \nspace{H}$ such that $\forall \epsilon > 0$
    \begin{equation*}
        ||\pmb{g} - \pmb{f}||_{\infty} = \underset{\pmb{x} \in \nspace{X}}{\sup}|\pmb{g}(\pmb{x}) - \pmb{f}(\pmb{x})| \leqslant \epsilon
    \end{equation*}
    
    \item \label{cond:A3_} The kernel functions $K_{1}^{(1)}(\cdot, \cdot)$, $K_{1}^{(2)}(\cdot, \cdot)$,  $K_{2}(\cdot, \cdot)$, and $K_{3}(\cdot, \cdot)$ used to composite the coupled tensor kernel (\ref{equ:C_TK_}) are all bounded, and are satisfying
    \begin{equation*}
        \sqrt{\sup K(\cdot, \cdot)} \leqslant K_{max} < \infty
    \end{equation*}
    for every kernel function mentioned above.
    
    \item \label{cond:A4_} The hyper-parameter in the regularization term $\lambda = \lambda_{n}$ satisfies:
    \begin{equation*}
        \begin{split}
             \lambda_{n} \rightarrow 0 \quad &\text{as} \quad  n \rightarrow \infty\\
             n\lambda_{n} \rightarrow \infty \quad &\text{as} \quad  n \rightarrow \infty\\
              \end{split}
    \end{equation*}
\end{enumerate}
\label{prop:consistent_}
\end{proposition}
\noindent This proposition is an extension of our previous result for the statistical consistency of CP-STM. The proof of this proposition is provided in Appendix \ref{append:proof_consistency_}. 

\section{Simulation Study}
\label{sec:simulation_}
We present a simulation study to demonstrate the benefit of utilizing C-STM with multimodal data in classification problems. To show the advantage of using multi-modalities in C-STM, we include CP-STM from \citet{he2014dusk}, Constrained Multilinear Discriminant Analysis (CMDA), and Direct General Tensor Discriminant Analysis (DGTDA) from \citet{li2014multilinear} as competitors. These existing approaches can only take a single tensor / matrix as the feature for classification. As a result, they are not able to enjoy the multi-modalities in the simulated data. We apply these approaches on every single data modality in our simulated data, and compare their classification performance with C-STM which uses multimodal data. 

We generate synthetic data using the idea from \citet{fanaee2016simtensor}. Suppose the two data modalities in our classification problems are
\begin{equation}
    \begin{split}
        & \ts{X}_{t, 1} = \sum \limits_{k = 1}^{3} \pmb{x}_{k, t, 1}^{(1)} \circ \pmb{x}_{k, t, 1}^{(2)} \circ \pmb{x}_{k, t, 1}^{(3)} \\ 
        & \pmb{X}_{t, 2} = \sum \limits_{k = 1}^{3} \pmb{x}_{k, t, 2}^{(1)} \circ \pmb{x}_{k, t, 2}^{(2)} 
    \end{split}
    \label{equ:simu_}
\end{equation}
where $\ts{X}_{t, 1}$ are three-way tensors in the size of 30 by 20 by 10. $\pmb{X}_{t, 2}$ are matrices in the size of 50 by 10. Both of them have CP ranks equal to 3. To generate data for the simulation study, we first generate the latent factors (vectors) from various multivariate normal distributions, and then convert these factors into full tensors $\ts{X}_{t, 1}$ and matrices $\pmb{X}_{t, 2}$ using equation (\ref{equ:simu_}). The multivariate normal distributions we used to generate the latent factors in equation (\ref{equ:simu_}) are specified in the table \ref{tab:simulation_setup} below. In the table \ref{tab:simulation_setup}, we use  $c = 1, 2$ to denote data from two different classes. 
\begin{table}[]
    \centering
    \begin{tabular}{l l c c c c}
    \toprule
    & & \multicolumn{2}{c}{Tensor Factors} & Shared Factors & Matrix Factors\\
    \midrule
     Simulation  & $c$ & $\pmb{x}_{k, t, 1}^{(1)}$ & $\pmb{x}_{k, t, 1}^{(2)}$ & $\pmb{x}_{k, t, 1}^{(3)} = \pmb{x}_{k, t, 2}^{(2)} $ & $\pmb{x}_{k, t, 2}^{(1)}$\\
     \midrule
      \multirow{2}{*}{Case 1} & 1 & $\pmb{MVN}(\pmb{1}, \pmb{I})$ & $\pmb{MVN}(\pmb{1}, \pmb{I})$ & $\pmb{MVN}(\pmb{1}, \pmb{I})$ & $\pmb{MVN}(\pmb{1}, \pmb{I})$ \\
      & 2 & $\pmb{MVN}(\pmb{1.5}, \pmb{I})$ & $\pmb{MVN}(\pmb{1}, \pmb{I})$ & $\pmb{MVN}(\pmb{1}, \pmb{I})$ & $\pmb{MVN}(\pmb{1.25}, \pmb{I})$ \\
     \midrule
     \multirow{2}{*}{Case 2} & 1 & $\pmb{MVN}(\pmb{1}, \pmb{I})$ & $\pmb{MVN}(\pmb{1}, \pmb{I})$ & $\pmb{MVN}(\pmb{1}, \pmb{I})$ & $\pmb{MVN}(\pmb{1}, \pmb{I})$ \\
      & 2 & $\pmb{MVN}(\pmb{1.5}, \pmb{I})$ & $\pmb{MVN}(\pmb{1}, \pmb{I})$ & $\pmb{MVN}(\pmb{1}, \pmb{I})$ & $\pmb{MVN}(\pmb{1.5}, \pmb{I})$ \\
      \midrule
      \multirow{2}{*}{Case 3} & 1 & $\pmb{MVN}(\pmb{1}, \pmb{I})$ & $\pmb{MVN}(\pmb{1}, \pmb{I})$ & $\pmb{MVN}(\pmb{1}, \pmb{I})$ & $\pmb{MVN}(\pmb{1}, \pmb{I})$ \\
      & 2 & $\pmb{MVN}(\pmb{1.5}, \pmb{I})$ & $\pmb{MVN}(\pmb{1}, \pmb{I})$ & $\pmb{MVN}(\pmb{1}, \pmb{I})$ & $\pmb{MVN}(\pmb{1.75}, \pmb{I})$ \\
      \midrule
      \multirow{2}{*}{Case 4} & 1 & $\pmb{MVN}(\pmb{1}, \pmb{I})$ & $\pmb{MVN}(\pmb{1}, \pmb{I})$ & $\pmb{MVN}(\pmb{1}, \pmb{I})$ & $\pmb{MVN}(\pmb{1}, \pmb{I})$ \\
      & 2 & $\pmb{MVN}(\pmb{1.5}, \pmb{I})$ & $\pmb{MVN}(\pmb{1}, \pmb{I})$ & $\pmb{MVN}(\pmb{1}, \pmb{I})$ & $\pmb{MVN}(\pmb{2}, \pmb{I})$ \\
      \midrule
      \multirow{2}{*}{Case 5} & 1 & $\pmb{MVN}(\pmb{1}, \pmb{I})$ & $\pmb{MVN}(\pmb{1}, \pmb{I})$ & $\pmb{MVN}(\pmb{1}, \pmb{I})$ & $\pmb{MVN}(\pmb{1}, \pmb{I})$ \\
      & 2 & $\pmb{MVN}(\pmb{1.5}, \pmb{I})$ & $\pmb{MVN}(\pmb{1}, \pmb{I})$ & $\pmb{MVN}(\pmb{1}, \pmb{I})$ & $\pmb{MVN}(\pmb{2.25}, \pmb{I})$ \\
      \midrule 
      \multirow{2}{*}{Case 6} & 1 & $\pmb{MVN}(\pmb{1}, \pmb{I})$ & $\pmb{MVN}(\pmb{1}, \pmb{I})$ & $\pmb{MVN}(\pmb{1}, \pmb{I})$ & $\pmb{MVN}(\pmb{1}, \pmb{I})$ \\
      & 2 & $\pmb{MVN}(\pmb{2}, \pmb{I})$ & $\pmb{MVN}(\pmb{1}, \pmb{I})$ & $\pmb{MVN}(\pmb{1}, \pmb{I})$ & $\pmb{MVN}(\pmb{1}, \pmb{I})$ \\
      \midrule
      \multirow{2}{*}{Case 7} & 1 & $\pmb{MVN}(\pmb{1}, \pmb{I})$ & $\pmb{MVN}(\pmb{1}, \pmb{I})$ & $\pmb{MVN}(\pmb{1}, \pmb{I})$ & $\pmb{MVN}(\pmb{1}, \pmb{I})$ \\
      & 2 & $\pmb{MVN}(\pmb{1}, \pmb{I})$ & $\pmb{MVN}(\pmb{1}, \pmb{I})$ & $\pmb{MVN}(\pmb{1}, \pmb{I})$ & $\pmb{MVN}(\pmb{2}, \pmb{I})$ \\
      \midrule
      \multirow{2}{*}{Case 8} & 1 & $\pmb{MVN}(\pmb{1}, \pmb{I})$ & $\pmb{MVN}(\pmb{1}, \pmb{I})$ & $\pmb{MVN}(\pmb{1}, \pmb{I})$ & $\pmb{MVN}(\pmb{1}, \pmb{I})$ \\
      & 2 & $\pmb{MVN}(\pmb{1}, \pmb{I})$ & $\pmb{MVN}(\pmb{1}, \pmb{I})$ & $\pmb{MVN}(\pmb{2}, \pmb{I})$ & $\pmb{MVN}(\pmb{1}, \pmb{I})$ \\
      \bottomrule
    \end{tabular}
    \caption{Distribution Specifications for Simulation Study; $\pmb{MVN}$ stands for multivariate normal distribution. $\pmb{I}$ are identity matrices. Bold numbers are vectors whose elements are all equal to the numbers.}
    \label{tab:simulation_setup}
\end{table}
There are eight different cases in our simulation study. In case 1 - 5, one of the tensor factors and the matrix factors are generated from different multivariate normal distributions for data in different classes. This means the tensor and matrix data both contain certain class information (discriminant power) which are different in different data modalities. Notice that the discriminant power in one of the tensor factor remains the same among case 1 - 5, while the power in the matrix factor increases. Case 6 and 7 assume the class information exists only in a single data modality. In case 6, only one of the tensor factors are generated from different distributions for data in different classes. This factor then becomes the matrix factor in class 7. In case 8, the shared factors are sampled from different distributions, meaning that both tensor and matrix data modalities have class information. However, such class information are from the shared factors are the same between different modalities.  

\begin{figure}
    \centering
    \includegraphics[width = \textwidth]{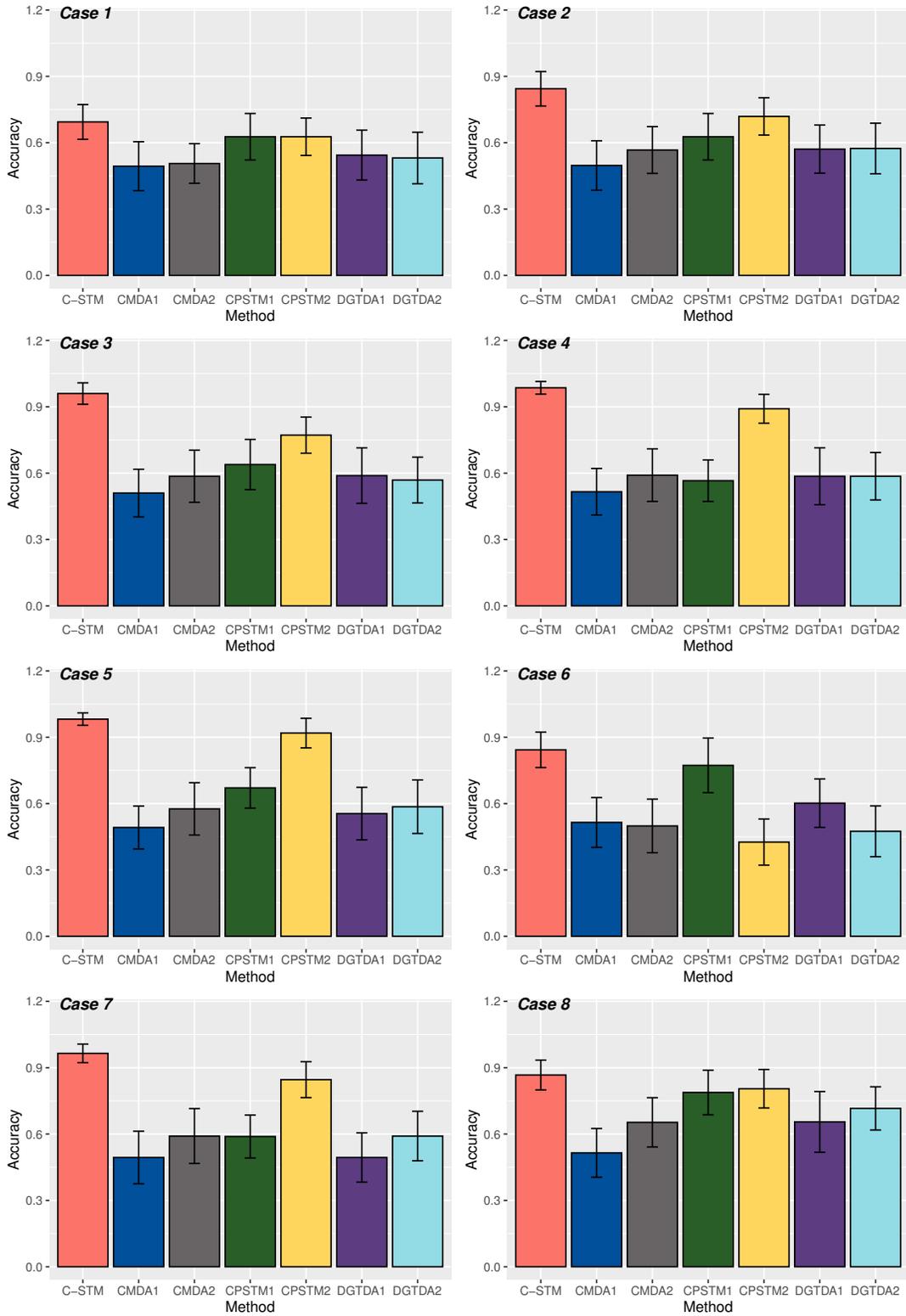}
    \caption{Simulation Result: Average accuracy rates shown in bar plots; Standard deviation of accuracy rates shown by error bars}
    \label{fig:simulation_result}
\end{figure}

For each simulation case, we generate 50 pairs of tensor and matrix from both class, collecting 100 pairs of observations in total. We then perform a random training and testing set separation by randomly choosing 20 samples as the testing set, and use the remaining data as the training set. The random selection of testing set is conducted in a stratified sampling manner such that the proportion of samples from each class remains the same in both training and testing sets. For all models, we report the model prediction accuracy, the proportion of correct predictions over total predictions, on the testing set as the performance metric. The random training and testing set separation is repeated for 50 times and the average prediction accuracy of these 50 repetitions for all the cases are reported in Figure \ref{fig:simulation_result}. Additionally, the standard deviations are illustrated by the error bars in the figure. The results of CP-STM, CMDA, and DGTDA with tensor data are denoted by CPSTM1, CMDA1, and DGTDA1 respectively in the figure. The results using matrix data are denoted by CPSTM2, CMDA2, and DGTDA2. 

From the Figure \ref{fig:simulation_result}, we can conclude that our C-STM has a more favorable performance in this multimodal classification problem comparing with other competitors. Its accuracy rates are significantly larger than other methods in most cases. Particularly, we can see that the accuracy rates of C-STM (pink) are increasing from case 1 to case 5, while the accuracy rates of CP-STM using tensor data remain the same. This is because the difference between class mean vectors for the first tensor factor does not change from case 1 to case 5. However, the gap between class mean vectors in matrix factor increases. Due to this fact, both C-STM and CP-STM (yellow) which utilize matrix data are getting better performance from case 1 to case 5. More importantly, C-STM always outperforms CP-STM with matrix data as it enjoys the extra class information from multimodalities. In case 6 and case 7 where class information are in single data modalities, the advantage of C-STM is not as significant as the previous cases, though its performance are slightly better than CP-STM. This indicates C-STM can provide robust classification results even though extra data modalities do not provide any other class information. In case 8 where the class information is from the share factors, C-STM turns out to better recover the share factors and provides significantly better classification accuracy. Through this simulation, we showed that C-STM has a clear advantage of using multimodal data in classification problems, and are robust to redundant data modalities.

\section{Trial Classification for Simultaneous EEG-fMRI Data}
\label{sec:realdata_}
In this section, we present the application of the proposed method on simultaneous EEG-fMRI data. The simultaneous electroencephalography (EEG) with functional magnetic resonance imaging (fMRI) is one of the most popular non-invasive multimodal brain imaging techniques to study human brain function. EEG records electrical activity from the scalp resulting from ionic current within the neurons of the brain. Its millisecond temporal resolution makes it possible to record event-related potentials that occur in response to visual, auditory and sensory stimuli  (\citet{teplan2002fundamentals, abreu2018eeg}). While EEG provides high temporal resolution, its spatial resolution is limited by the number of electrodes placed on the scalp and thus provides less spatial resolution compared to other neuroimaging modalities such as  magnetic resonance imaging (MRI) and Positron Emission Tomography (PET).  As a result, it has been commonplace to record EEG data in conjunction with a high spatial resolution modality. As another powerful tool in studying human brain function,  blood oxygenation level dependent (BOLD) functional magnetic resonance imaging (fMRI) provides signals with much higher spatial resolution to reflect hemodynamic changes in blood oxygenation level at all voxels related to neuronal activities (\citet{ogawa1990brain, belliveau1991functional, kwong1992dynamic, filippi2005mr}). Recording simultaneous EEG and fMRI can provide high resolution information at both the spatial and temporal dimensions at the same time. Thus, developing novel machine learning techniques to utilize such multimodal data is of great significance. In this application, we apply our C-STM model to a binary trial classification problem on a simultaneous EEG-fMRI data. 

The data is obtained from the study \citet{walz2013simultaneous}. In this study, there are seventeen individuals (six females, average age 27.7) participated in three runs each of analogous visual and auditory oddball paradigms. The 375 (125 per run) total stimuli per task were presented for 200 ms each with a 2-3 s uniformly distributed variable inter-trial interval. A trial is defined as a time window in which subjects receive stimuli and make responses. In the visual task, a large red circle on isoluminant gray backgrounds was considered as the target stimuli, and a small green circle were the standard stimuli. For the auditory task, the standard and oddball stimuli were, respectively, 390 Hz pure tones and broadband sounds which sound like "laser guns". During the experiment, the stimuli were presented to all subjects, and their EEG and fMRI data are collected simultaneously and continuously. We obtain the EEG and fMRI data from OpenNeuro website (\url{https://openneuro.org/datasets/ds000116/versions/00003}). We utilize both EEG and fMRI in this data set with our C-STM model to class stimulus types in all the trials.  Through our numerical study, we want to demonstrate the fact our C-STM model enjoys the advantage of data multimodality and provides more accurate class predictions. The data from Subject 4 are dropped since its fMRI data are corrupted. Due to the fact that the number of trials from each subject are different, we further provide a table in appendix \ref{append:realdata_} to show the number of trials for each subject. 

\begin{figure}
     \centering
     \begin{subfigure}[b]{\textwidth}
         \centering
         \includegraphics[width=\textwidth]{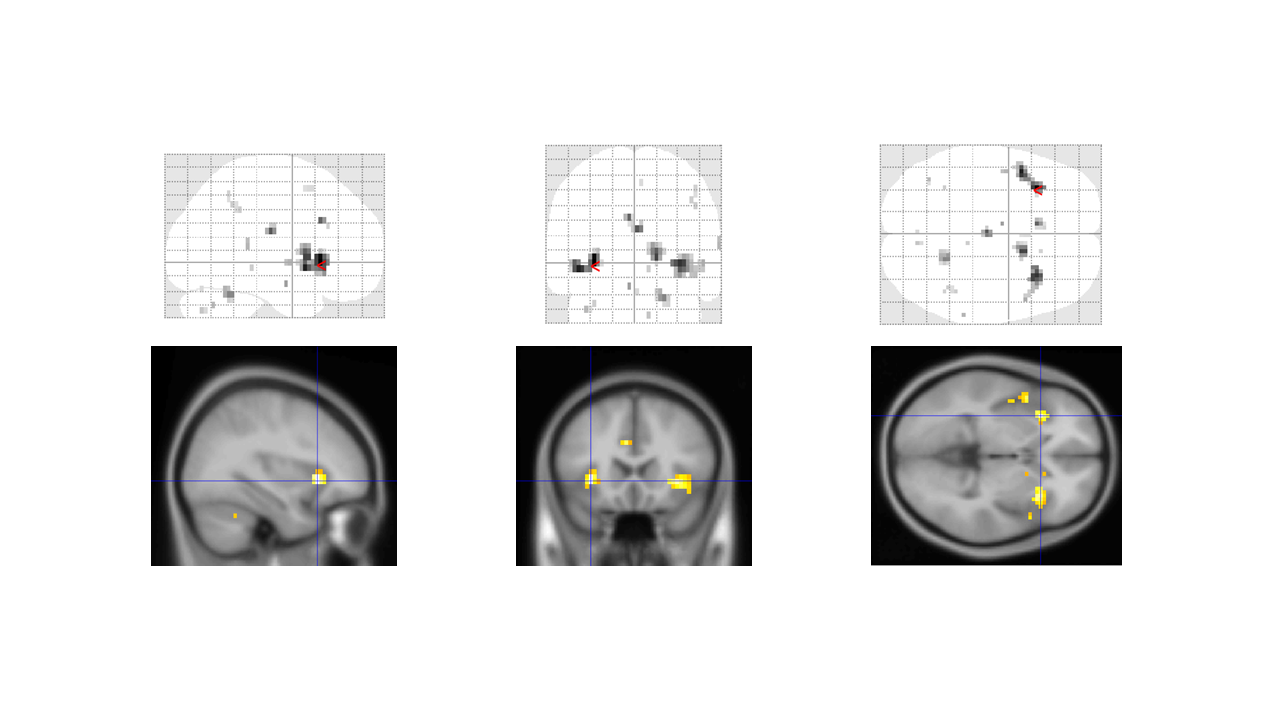}
         \caption{Auditory Task}
         \label{fig:roi_auditory_}
     \end{subfigure}
     \hfill
     \begin{subfigure}[b]{\textwidth}
         \centering
         \includegraphics[width=\textwidth]{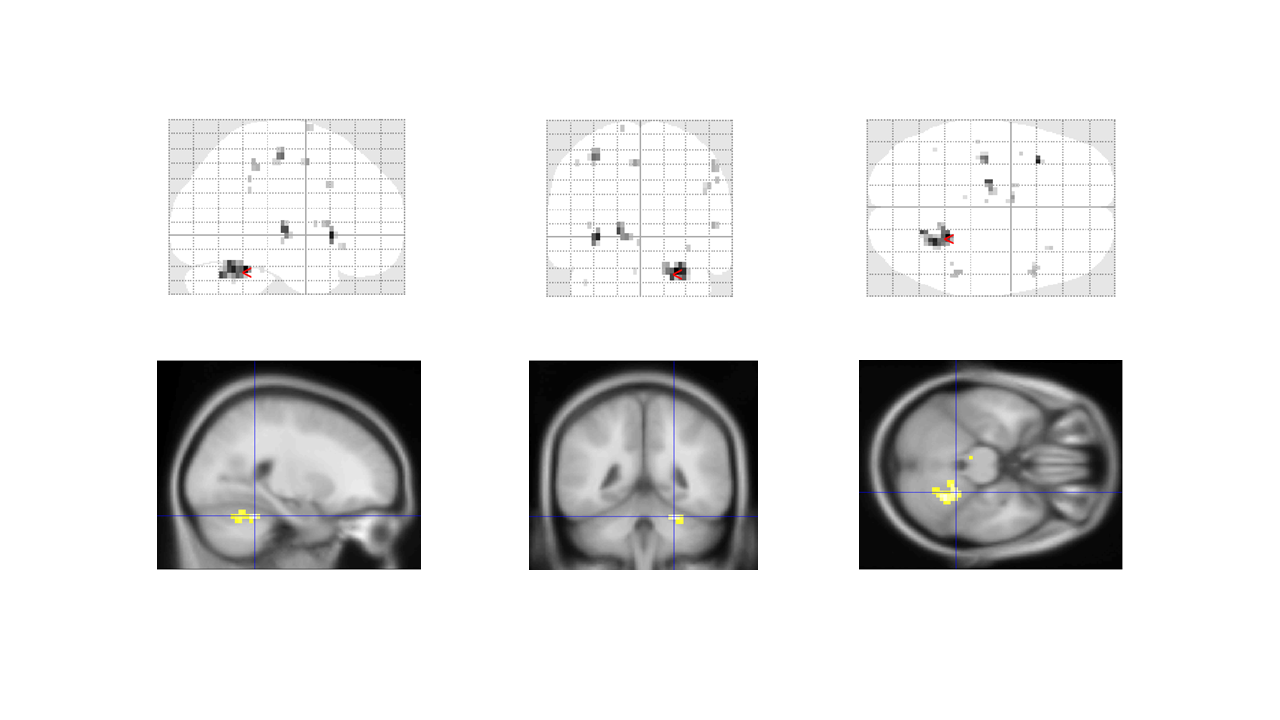}
         \caption{Visual Task}
         \label{fig:roi_visual_}
     \end{subfigure}
        \caption{Region of Interest (ROI)}
        \label{fig:roi_}
\end{figure}

We pre-process both the EEG and fMRI data with Statistical Parametric Mapping (SPM 12) (\citet{ashburner2014spm12}) and Matlab. The EEG data is collected by a custom built MR-compatible EEG system with 49 channels. \citet{walz2013simultaneous} provides a version of re-referenced EEG data with 34 channels which are used in our experiment. This version of EEG data are sampled at 1,000 Hz, and are downsampled to 200 Hz at the beginning of pre-processing. We then  remove both low-frequency and high-frequency noise in the data using SPM filter functions. As the last step of EEG pre-processing, we define trials from Brain Imaging Data Structure (BIDS) files \citet{niso2018meg} and extract EEG data epochs recorded within the trial-related time windows.  The time window for each trial is considered to go from 100 ms before the stimulus onset until 500 ms after the stimulus. For each trial, we construct a three-mode tensor corresponding to the EEG data for all subjects where the modes represent channel $\times$ time $\times$ subject. We denote it as $\ts{X}_{t, 1} \in \mathbb{R}^{34 \times 121 \times 16}$. The fMRI data is collected by 3T Philips Achieva MR Scanner with 170 volumes (TR = 2s) per session. Each 3D volume contains 32 slices. The voxel size in the image is 3 x 3 x 4 mm. For each subject, we realign all the fMRI volumes from multiple sessions to the mean volume, and co-register the participant's T1 weighted  anatomical scan to the mean fMRI volume. Next, we normalize all the fMRI volumes to match the MNI brain template (\citet{lancaster2007bias}) by creating segments from co-registered T1 weighted scan, and keep the voxel size as 3 x 3 x 4 mm. All normalized fMRI volumes are then smoothed by 3D Gaussian kernels with full width at half maximum (FWHM) parameter being $8 \times 8 \times 8$. After the pre-processing, we further perform a regular statistical analysis (\citet{lindquist2008statistical, worsley2002general}) to extract fMRI volumes from visual and auditory stimulus related voxels. Such data are also known as Region of Interest (ROI) data. We describe the ROI voxel identification and data extraction in Appendix \ref{append:realdata_}. We extract fMRI volmes from 178 voxels (in Figure \ref{fig:roi_auditory_}) for auditory oddball tasks, and 112 voxels for auditory tasks. As a result, fMRI data are modeled by matrices whose rows and columns stand for voxels and subjects: $\ts{X}_{t, 2} \in \mathbb{R}^{16 \times 178}$ for auditory task data, and $\ts{X}_{t, 2} \in \mathbb{R}^{16 \times 112}$ for visual task data. There is no time mode in fMRI data because the trial duration is less than the repetition time of fMRI (time for obtaining a single 3D volume fMRI). For each trial, there is only one 3D scan of fMRI collected from a single subject. The ROI data then becomes a vector for this subject in the trial as we extract volumes from the regions of interest.

To classify trials with oddball and standard stimulus, we collect 140 multimodal data samples $(\ts{X}_{t, 1}, \ts{X}_{t, 2})$ from auditory tasks, and 100 samples from visual tasks. For both types of tasks, the numbers of oddball and standard trials are equal. We consider the trials with oddball stimulus as the positive class, and the trials with standard stimulus as the negative class. Like the procedures in our simulation study, we select 20\% of data as testing set, and use the remaining 80\% for model estimation and validation. The classification accuracy, precision (positive predictive rate), sensitivity (true positive rate), and specificity (true negative rate) of classifiers are calculated using the test set at each experiment. The experiment is repeated multiple times, and the average accuracy, precision, sensitivity, and specificity, and their standard deviations (in subscripts) are reported in Table \ref{tab:eeg_fMRI_result_}. The single mode classifiers CPSTM, CMDA, and DGTDA are also applied on either EEG or fMRI data as a comparison. The single mode classifiers applied on EEG data are denoted by appending the number "1" after their names, and those applied on fMRI data are denoted by appending the number "2". The area under the curve (AUC) for all the classifiers are also reported in Table \ref{tab:eeg_fMRI_result_}. 

\begin{table}[]
    \centering
    \begin{tabular}{c l c c c c c c}
    \toprule
     Task & Method & Accuracy & Precision & Sensitivity & Specificity & AUC \\
     \midrule
   \multirow{7}{*}{Auditory} & C-STM & $\pmb{0.89}_{0.05}$ & $0.83_{0.07}$ & $1.00_{0.00}$ &  $0.77_{0.11}$ & $\pmb{0.89}_{0.06}$\\
                               & CP-STM1 & $0.80_{0.08}$ & $0.71 _{0.11}$ & $1.00 _{0.00}$ & $0.60_{0.12}$ & $0.78_{0.06}$\\
                               & CP-STM2 & $0.83_{0.06}$ & $0.76_{0.07}$ & $0.99_{0.05}$ & $0.65_{0.11}$ & $0.82_{0.05}$\\
                               & CDMA1 & $0.55_{0.10}$ & $0.51_{0.09}$ & $0.96_{0.09}$ & $0.20  _{0.21}$ & $0.55_{0.06}$\\
                               & CDMA2 & $0.67_{0.09}$ & $0.61_{0.11}$ & $0.92_{0.07}$ & $0.46_{0.14}$ & $0.70_{0.08}$\\
                               & DGTDA1 & $0.55_{0.09}$ & $0.51_{0.09}$ & $0.94_{0.07}$ & $0.23_{0.12}$ & $0.59_{0.06}$\\
                               & DGTDA2 & $0.67_{0.09}$ & $0.60_{0.10}$ & $0.90_{0.09}$ & $0.46_{0.13}$ & $0.68_{0.08}$\\
     \midrule
    \multirow{7}{*}{Visual} & C-STM & $\pmb{0.86}_{0.06}$ & $0.82_{0.09}$ & $0.93_{0.07}$ &  $0.77_{0.12}$ & $\pmb{0.86}_{0.06}$\\
                               & CP-STM1 & $0.76_{0.08}$ & $0.66 _{0.11}$ & $1.00 _{0.00}$ & $0.54_{0.12}$ & $0.78_{0.05}$\\
                               & CP-STM2 & $0.77_{0.08}$ & $0.70_{0.11}$ & $0.98 _{0.08}$ & $0.58_{0.17}$ & $0.77_{0.07}$\\
                               & CDMA1 & $0.53_{0.12}$ & $0.52_{0.11}$ & $0.94_{0.11}$ & $ 0.11_{0.18}$ & $0.54_{0.08}$\\
                               & CDMA2 & $0.65_{0.13}$ & $0.61_{0.14}$ & $0.91_{0.09}$ & $0.43_{0.19}$ & $0.66_{0.09}$\\
                               & DGTDA1 & $0.56_{0.11}$ & $0.54_{0.11}$ & $0.94_{0.06}$ & $0.17_{0.12}$ & $0.56_{0.07}$\\
                               & DGTDA2 & $0.64_{0.10}$ & $0.60_{0.13}$ & $0.86_{0.10}$ & $0.44_{0.18}$ & $0.64_{0.07}$\\
     \bottomrule
    \end{tabular}
    \caption{Real Data Result: Simultaneous EEG-fMRI Data Trial Classification (Mean of Performance Metrics with Standard Deviations in Subscripts)}
    \label{tab:eeg_fMRI_result_}
\end{table}
The results in Table \ref{tab:eeg_fMRI_result_} show that the trial classification accuracy for C-STM using multimodal data is better than any classifier based on single modality with a significant improvement in terms of average accuracy rates and average AUC values. This improvement is observed for classification of  both auditory and visual tasks. This observation agrees to the conclusion from our simulation study. Similar to our simulation study, the tensor discriminant analysis does not work as well as CP-STM and C-STM. In addition, it is obvious that the performance of tensor discriminant analysis using fMRI data are better than using EEG data. This is within our expectation, since the regions we extracted from fMRI data are identified by group level fMRI statistical analysis (see Appendix \ref{append:realdata_}). The data in these regions have already shown significant differences between different trials in the traditional study, and thus are easy to classify. On the other hand, there is no prior analysis and feature extraction procedure applied on EEG data, leaving a low signal to noise ratio in EEG data. However, C-STM still can take advantage of using EEG data and further increase the classification accuracy, highlighting its robustness and potential in processing noisy mulitmodal tensor data.

\section{Conclusion}
\label{sec:conclusion_}
In this work, we have proposed a novel coupled support tensor machine classifier for multimodal data  by combining the advanced coupled matrix tensor factorization (ACMTF) and support tensor machine (STM). The most distinctive feature of this classifier is its ability to integrate features across different modalities and structures.  The proposed approach can simultaneously take  matrix- and tensor-shaped data for classification and can be easily extended to inputs with more than two modality. The coupled tensor matrix decomposition unveils the intrinsic correlation structure between data across different modalities, making it possible to integrate information from multiple sources efficiently. Such a decomposition also makes the whole method robust and applicable to large-scale noisy data with missing values.
The newly designed kernel functions in C-STM provide feature-level information fusion, combining discriminant information from different modalities. Moreover, the kernel formulation makes it possible to utilize the most discriminative features from each modality by tuning the weight parameters in the function.  Our theoretical results demonstrate that the C-STM decision rule is statistically consistent. 

The most important theoretical extension of our current approach would be the development of excess risk for C-STM. In particular, we are looking for an explicit expression for the excess risk in terms of data factors from multiple modalities to quantify the contribution of every single modality in minimizing the excess risk. By doing so, we are able to interpret the importance of each data modality in classification tasks. In addition, quantifying the uncertainty of tensor and matrix factors estimation and their impact on the excess risk will build the foundation to the next level. 

Future work will focus on learning the weight parameters in the kernel function via optimization. As \citet{gonen2011multiple} introduced, the weights in the kernel function can be further estimated by including a group lasso penalty in the objective function. Such a weight estimation procedure can identify the most significant data components and reduce the burden of parameter selection. In addition, the proposed framework can be extended to multimodal tensors with more than two modalities, and for regression problems. 

In conclusion, we believe C-STM offers many encouraging possibilities for multimodal data integration and analysis. Its capability of handling multimodal tensor inputs will make it appropriate in many advanced data applications in neuroscience and medical research.  We anticipate that this method will play an important role in a variety of applications.

\newpage

\appendix
\section{Proof of Theorem \ref{prop:consistent_}}
\label{append:proof_consistency_}
\begin{proof}
To prove the proposition \ref{prop:consistent_}, we introduce few more notations here. Let $\mathcal{L}$ be the loss function satisfying the condition \ref{cond:A2_}. We denote the classification risk for an arbitrary decision function, $\pmb{f}$, as 
\begin{equation*}
       \mathcal{R}_{\mathcal{L}}(\pmb{f}) = \mathbb{E}_{\mathcal{X} \times \mathcal{Y}} \mathcal{L}(y, \pmb{f}(\ts{X})) = \int \mathcal{L}(y, \pmb{f}(\ts{X})) d\mathbb{P}
\end{equation*}
The expectation is taken over the joint distribution of $\nspace{X} \times \nspace{Y}$. Notice that this risk notation, $\mathcal{R}_{\mathcal{L}}(\pmb{f})$, is different from our notation $\mathcal{R}(\pmb{f})$ in section \ref{sec:theory_} since we use the Lipschitz continuous loss $\mathcal{L}$ instead of the "zero-one" loss to measure the classification error. $\mathcal{L}$ is also called surrogate loss for classification problems. Examples of such surrogate loss functions include Hinge loss and Squared Hinge loss. Comparison of these loss functions and their statistical properties can be found in \citet{zhang2004statistical}. If we denote the Bayes risk under the surrogate loss $\mathcal{L}$ as $\mathcal{R}_{\mathcal{L}}^{*}$, i.e. $\mathcal{R}_{\mathcal{L}}^{*} = \min \mathcal{R}_{\mathcal{L}}(\pmb{f})$ for all measurable function $f$, then the result from \citet{zhang2004statistical} says $\mathcal{R}_{\mathcal{L}}(\pmb{f_{n}}) \rightarrow \mathcal{R}_{\mathcal{L}}^{*}$ indicates  $\mathcal{R}(\pmb{f_{n}}) \rightarrow \mathcal{R}^{*}$ for any decision rule $\{ \pmb{f_{n}}\}$. This conclusion holds as long as the surrogate loss is "self-calibrated" (see \citet{steinwart2008support}). Since we use Hinge loss in our problem, and Hinge loss is known to be Lipschitz and self-calibrated, our assumption \ref{cond:A2_} holds in our discussion. Thus, we only need to show $\mathcal{R}_{\mathcal{L}}(\pmb{f_{n}}) \rightarrow \mathcal{R}_{\mathcal{L}}^{*}$ for the proof of our proposition \ref{prop:consistent_}.

Given the tuning parameter $\lambda$ satisfying condition \ref{cond:A4_}, we denote 
\begin{equation*}
    \pmb{f_{n}^{\lambda}} = \underset{\pmb{f} \in \nspace{H}}{\arg \min} \quad \lambda \cdot ||\pmb{f}||^{2} + \frac{1}{n} \sum \limits_{i = 1}^{n}\mathcal{L}(\pmb{f}(\ts{X}_{i}), y_{i})
\end{equation*}
where $\nspace{H}$ is the reproducing kernel Hilbert space (RKHS) generated by the kernel function (\ref{equ:C_TK_}). As we mentioned in the section \ref{sec:methodology_}, $\nspace{H}$ is also know as the collection of functions which are in the form of equation (\ref{equ:C_STM_decision_}). Now we further assume 
\begin{equation*}
    \pmb{f^{\lambda}} = \underset{\pmb{f} \in \nspace{H}}{\arg \min} \quad \lambda \cdot ||\pmb{f}||^{2} + \mathcal{R}_{\mathcal{L}}(\pmb{f})
\end{equation*}
Then $\pmb{f^{\lambda}}$ is the optimal decision function from $\nspace{H}$ such that it minimizes the expected risk. Comparing $\pmb{f_{n}^{\lambda}}$ with $\pmb{f^{\lambda}}$, we can understand that $\pmb{f^{\lambda}}$ is the version of $\pmb{f_{n}^{\lambda}}$ when the size of training data is as large as possible. If we denote $\mathcal{R}_{\mathcal{L}, T_{n}} (\pmb{f}) = \frac{1}{n} \sum \limits_{i = 1}^{n}\mathcal{L}(\pmb{f}(\ts{X}_{i}), y_{i})$, then $\mathcal{R}_{\mathcal{L}, T_{n}} (\pmb{f})$ is a sample estimate of $\mathcal{R}_{\mathcal{L}}(\pmb{f})$. With $\pmb{f^{\lambda}}$, we can show that 
\begin{equation*}
     |\mathcal{R}_{\mathcal{L}}(\pmb{f^{\lambda}_{n}}) - \mathcal{R}_{\mathcal{L}}^{*}| \leqslant |\mathcal{R}_{\mathcal{L}}(\pmb{f_{n}^{\lambda}}) - \mathcal{R}_{\mathcal{L}}(\pmb{f^{\lambda}})| + |\mathcal{R}_{\mathcal{L}}(\pmb{f^{\lambda}}) - \mathcal{R}_{\mathcal{L}}^{*}|
\end{equation*}
through triangular inequality. Since the Bayes risk under loss function $\mathcal{L}$ is defined as $\mathcal{R}^{*} = \underset{\pmb{f}: \nspace{X} \rightarrow \nspace{Y}}{\min} \mathcal{R}(\pmb{f})$ over all functions defined on $\nspace{X}$, we can immediate show that 
\begin{equation}
    \begin{split}
        |\mathcal{R}(\pmb{f^{\lambda}}) - \mathcal{R}^{*}| & \leqslant \mathbb{E}_{(\mathcal{X} \times \mathcal{Y}) } |\mathcal{L}(y, \pmb{f^{\lambda}}(\ts{X})) - \mathcal{L}(y, \pmb{f^{*}}(\ts{X}))| \leqslant C(K_{max}) \sup|\pmb{f^{\lambda}} - \pmb{f^{*}}| \\
        & \leqslant C(K_{max}) \cdot \epsilon
    \end{split}
    \label{equ_error_decom_append}
\end{equation}
This is the result of using condition \ref{cond:A1_} and \ref{cond:A2_} in the proposition \ref{prop:consistent_}. $\pmb{f^{\lambda}}$ is in the RKHS and thus bounded by some constant depending on $K_{max}$. $\pmb{f^{*}}$ is also continuous on compact subspace $\nspace{X}$ (because all the tensor components considered are bounded in condition \ref{cond:A1_}) and thus is bounded. The universal approximating property in condition \ref{cond:A3_} makes equation (\ref{equ_error_decom_append}) vanishes as $\epsilon$ goes to zero. Thus, the consistency result can be established if we show  $ |\mathcal{R}(\pmb{f_{n}^{\lambda}}) - \mathcal{R}(\pmb{f^{\lambda}})|$ converges to zero. This can be done with Rademacher complexity (see Chapter 26 in \citet{shalev2014understanding}).  

From the objective function (\ref{equ:C-STM_obj_}), we have 
\begin{equation}
    \mathcal{R}_{\mathcal{L}, T_n}(\pmb{f_{n}}) + \lambda_{n} ||\pmb{f_{n}}||^{2} \leqslant L_{0}
\end{equation}
under condition \ref{cond:A2_} when we simply let $\pmb{f} = 0$ as a naive classifier. Thus, $||\pmb{f_{n}}|| \leqslant \sqrt{\frac{L_{0}}{\lambda_{n}}}$. Let $M_{n} = \sqrt{\frac{L_{0}}{\lambda_{n}}}$. $\pmb{f_{\epsilon}} \in \nspace{H}$ such that $\mathcal{R}_{\mathcal{L}}(\pmb{f_{\epsilon}}) \leqslant \mathcal{R}_{\mathcal{L}}(\pmb{f^{\lambda}}) + \frac{\epsilon}{2}$. $ ||\pmb{f_{\epsilon}}|| \leqslant M_{n}$ when $n$ is sufficiently large.  Due to condition \ref{cond:A4_}, $\lambda_{n} \rightarrow 0$, making $M_{n} \rightarrow \infty$. Further notice that we introduce $\pmb{f_{\epsilon}}$ since it is independent of $n$. As a result, its norm, even though is bounded by $M_{n}$, is a constant and is not changing with respect to $n$.
By Rademacher complexity, the following inequality holds with probability at least $1 - \delta$, where $0 < \delta < 1$
 \begin{align*}
        && \mathcal{R}_{\mathcal{L}}(\pmb{f_{n}^{\lambda}})  & \leqslant \mathcal{R}_{\mathcal{L}, T_n}(\pmb{f_{n}^{\lambda}}) + \frac{2 C(K_{max}) M_{n}}{\sqrt{n}} + (L_{0} + C(K_{max}) M_{n}) \sqrt{\frac{\log 2 / \delta}{2n}}\\
        &&\text{\tiny $\pmb{f_{\epsilon}}$ is not the optimal in training data} \quad & \leqslant \mathcal{R}_{\mathcal{L}, T_n}(\pmb{f_{\epsilon}}) + \lambda_{n}||\pmb{f_{\epsilon}}||^{2} - \lambda_{n}||\pmb{f_{n}^{\lambda}}||^{2} + \frac{2 C(K_{max}) M_{n}}{\sqrt{n}} \\
        && & + (L_{0} + C(K_{max}) M_{n}) \sqrt{\frac{\log 2 / \delta}{2n}}\\
        &&\text{\tiny Drop ($\lambda_{n}||\pmb{f_{n}^{\lambda}}||^{2} > 0$) } \quad & \leqslant  \mathcal{R}_{\mathcal{L}, T_n}(\pmb{f_{\epsilon}})  + \lambda_{n}||\pmb{f_{\epsilon}}||^{2} + \frac{2 C(K_{max}) M_{n}}{\sqrt{n}} \\
        && & + (L_{0} + C(K_{max}) M_{n}) \sqrt{\frac{\log 2 / \delta}{2n}}\\
        &&\text{\tiny Rademacher Complexity again } \quad & \leqslant  \mathcal{R}_{\mathcal{L}}(\pmb{f_{\epsilon}})  + \lambda_{n}||\pmb{f_{\epsilon}}||^{2} + \frac{4 C(K_{max}) M_{n}}{\sqrt{n}} \\
        && & + 2(L_{0} + C(K_{max}) M_{n}) \sqrt{\frac{\log 2 / \delta}{2n}}\\
\end{align*}
Let $\delta = \frac{1}{n^{2}}$, and $N$ large such that for all $n > N$, 
\begin{equation*}
    \begin{split}
        \lambda_{n}||\pmb{f_{\epsilon}}||^{2} + \frac{4 C(K_{max}) M_{n}}{\sqrt{n}}  + 2(L_{0} + C(K_{max}) M_{n}) \sqrt{\frac{\log 2 / \delta}{2n}} \leqslant \frac{\epsilon}{2}
    \end{split}
\end{equation*}
The inequality exists because $||\pmb{f_{\epsilon}}||$ is a constant with respect to $n$, and all other terms are converging to zero. Thus
$$\mathcal{R}_{\mathcal{L}}(\pmb{f_{n}^{\lambda}}) \leqslant \mathcal{R}_{\mathcal{L}}(\pmb{f_{\epsilon}}) +  \frac{\epsilon}{2} \leqslant \mathcal{R}_{\mathcal{L}}(\pmb{f^{\lambda}}) + \epsilon$$
with probability $1 - \frac{1}{n^{2}}$. We conclude that 
\begin{equation}
    \mathbb{P}( |\mathcal{R}_{\mathcal{L}}(\pmb{f_{n}^{\lambda}}) - \mathcal{R}_{\mathcal{L}}(\pmb{f^{\lambda}}) | \geqslant \epsilon) \rightarrow 0
\end{equation}
for any arbitrary $\epsilon$. This establishes the weak consistency of CP-STM. For strong consistency, we consider for each $n$
$$\sum \limits_{n = 1}^{\infty}  \mathbb{P}( |\mathcal{R}_{\mathcal{L}}(\pmb{f_{n}^{\lambda}}) - \mathcal{R}_{\mathcal{L}}(\pmb{f^{\lambda}}) | \geqslant \epsilon) \leqslant N - 1 + \sum \limits_{n = 1}^{\infty}\frac{1}{n^{2}} \leqslant \infty$$
By Borel-Cantelli Lemma (\cite{durrett2019probability}), $\mathcal{R}_{\mathcal{L}}(\pmb{f_{n}^{\lambda}}) \rightarrow \mathcal{R}_{\mathcal{L}}(\pmb{f^{\lambda}})$ almost surely. The proof is finished.
\end{proof}

\section{Data Pre-processing for Section \ref{sec:realdata_}}
\label{append:realdata_}
We provide further details about our EEG-fMRI data pre-processing and fMRI data extraction in this section. Most of the processing steps are referred from \cite{henson2019multimodal}.

\subsection{fMRI Data}
The fMRI data processing includes three major steps, which are pre-processing, regions of interests (ROI) identification, and data extraction. We describe all these steps here. All the steps are performed by SPM 12 in Matlab. There are five steps in the image pre-processing part including realignment, co-registration, segment, normalization, and smoothing. 
\begin{itemize}
    \item \textbf{Realignment}: It is a procedure to align all the 3D BOLD volumes recorded along the time to remove artifacts caused by head motions, and also to estimate head position. For each task, there are three sessions of fMRI scans, providing 510 scans in total for each subject. These scans are realigned within subject to the average of these 510 scans. (average across time) In SPM, we create three independent sessions to load all the fMRI runs, and choose not to reslice all the images at this step. The reslicing will be done in normalization step. Avoiding extra reslicing can avoid introducing new artifacts. The mean scan is created in this step for co-registration. 
    
    \item \textbf{Co-registration}: Since all the fMRI scans are aligned to the mean scan, we have to transform the T1 weighted anatomical scan to match their orientation. Reason for doing this is that all the data will finally be transformed to a standardized space. Estimating such a transformation with T1 weighted scan can provide a high accuracy, since anatomical scans have higher resolutions. Matching the orientation of T1 weighted scan with all the fMRI scans makes it possible to apply the transformation estimated from T1 scan directly on fMRI data. In this step, we let the mean fMRI scan to be stationary, and move T1 anatomical scan to match it. A resliced T1 weighted scan is created in this step.
    
    \item \textbf{Segment}: This step estimate a deformation transformation mapping data into MNI 152 template space \cite{lancaster2007bias, brett2001using}. A forward deformation field is created in this step. 
    
    \item \textbf{Normalization}: In this step, the forward deformation is applied to all realigend fMRI scans, transforming all the data into MNI template space. The voxel size is set to be $3 \times 3 \times 4$ mm, which is the same as the original images. 
    
    \item \textbf{Smoothing}: All normalized fMRI volumes are then smoothed by 3D Gaussian kernels with full width at half maximum (FWHM) parameter being $8 \times 8 \times 8$.
\end{itemize}
This pre-processing procedure is applied to auditory and visual fMRI scans separately and independently. 

\begin{figure}[htb]
    \centering
    \includegraphics[height = \textwidth]{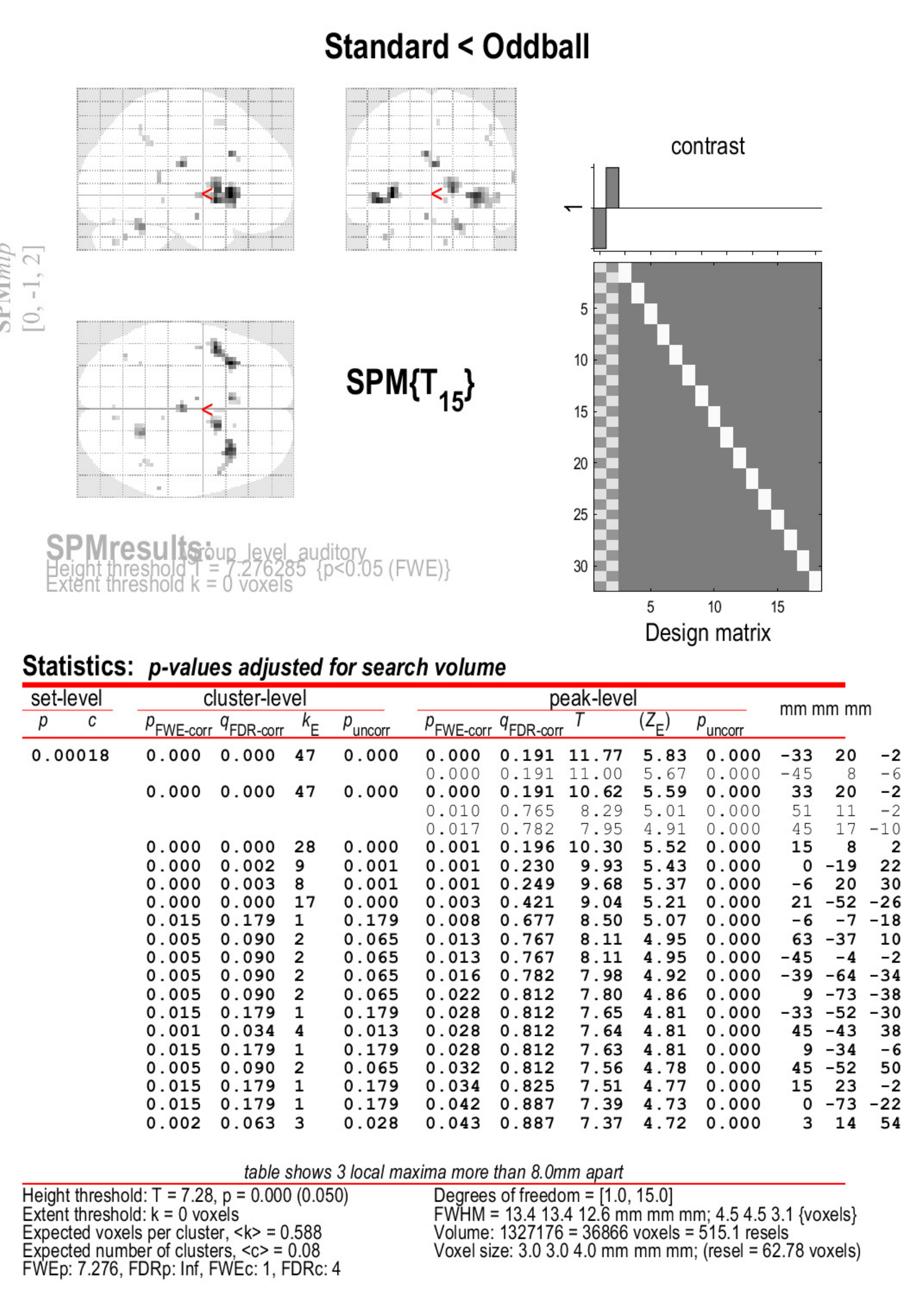}
    \caption{Auditory fMRI Group Level Analysis}
    \label{fig:auditory_level2}
\end{figure}

\begin{figure}[htb]
    \centering
    \includegraphics[height = \textwidth]{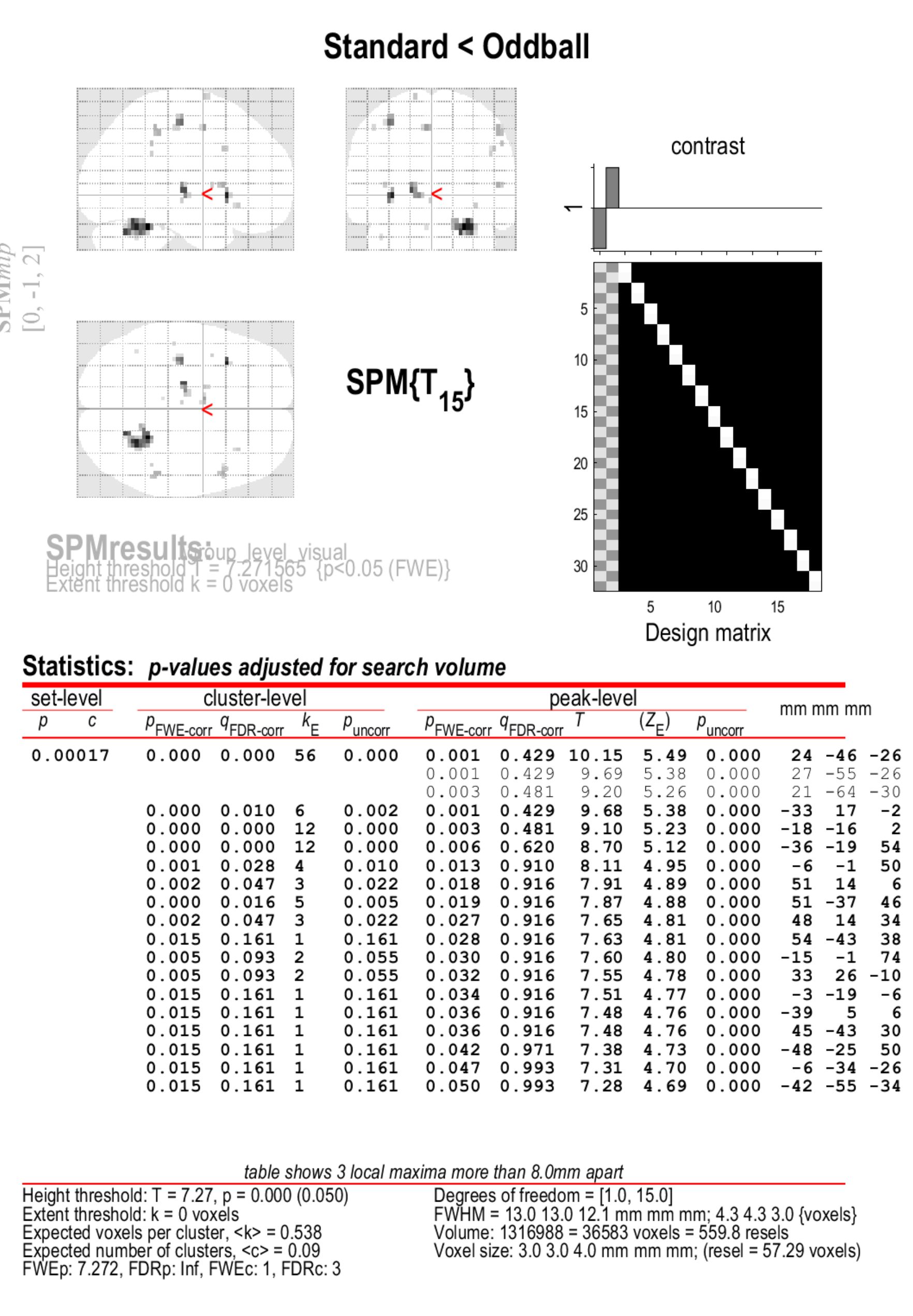}
    \caption{Visual fMRI Group Level Analysis}
    \label{fig:visual_level2}
\end{figure}

For each task, the processed fMRI are used to for statistcal analysis introduced in \cite{lindquist2008statistical, worsley2002general}. These models are basic linear mixed effect model with auto-regression covariance structure. Since these models are standard and are out of the scope of this dissertation, we do not introduce them in this part. For the first level (subject level) analysis, we use the model to estimate two contrast images: standard stimulus over baseline and oddball stimulus over baseline. These two are difference of average BOLD signals during stimulus time and that during no stimulus (baseline) time. They can be understand as the estimate $\hat{\beta}$ in a regression model $y = x\beta + \epsilon$. These contrasts are then pooled together in the group-level analysis. For each voxel, the group-level analysis performs a T-test to compare the BOLD signals in standard contrasts and oddball contrasts. For voxels whose test results is significant, SPM highlighted them as the regions of interest (ROI). The ROI of auditory and visual tasks are presented in the figure \ref{fig:auditory_level2} and figure \ref{fig:visual_level2} with P-values. 

The coordinates of these activate voxels are also provided in the statistical analysis results. To extract ROI data, we can use "spm\_get\_data" function in SPM 12. Since we are classifying trials, we only take one fMRI scan for each trial. This is because the trial duration (0.6 sec) is less than the repetition time (2 sec) of fMRI data. For each trial, we take the $k$-th fMRI scan where "k = round(onset / TR) + 1". This option is also inspired by SPM codes. 

\subsection{EEG Data}
The pre-processing of EEG data is relatively easy comparing to fMRI, since all EEG are already converted to MAT file and are re-referenced. Thus, we only need to resample it using Matlab Signal Processing toolbox to a lower sampling rates, which is 200 Hz in our case. Then, we use function "ft\_preproc\_lowpassfilter" and "ft\_preproc\_highpassfilter" from SPM 12 toolbox to filter the data. Finally, we split EEG into epochs which starts 100 ms before the onset and ends 500 ms after the onset. According to \cite{henson2019multimodal}, such a duration is long enough to capture the event-related potential for EEG data.

\begin{table}[htb]
    \centering
    \begin{tabular}{l c c c c}
    \toprule
      Tasks   & Auditory Oddball & Auditory Standard & Visual Oddball & Visual Standard \\
    \midrule
       Subject 1  & 75 &299 &75 &299\\
       Subject 2   & 70 &287 &70 &287\\
       Subject 3   & 74 & 296 &74 &296\\
       Subject 5  & 74 & 299 &74 &299\\
       Subject 6   & 75 & 290 &75 &290\\
       Subject 7  & 73 & 295 & 73 &295\\
       Subject 8   & 72 & 297 & 72 &297\\
       Subject 9   & 75 & 297 & 75 &298\\
       Subject 10  & 72 & 298 &72 &298\\
       Subject 11  & 70 & 293 & 70 &293\\
       Subject 12  & 74 & 299 &74 &299\\
       Subject 13   & 71 & 297 &71 &297\\
       Subject 14   & 75 & 296 & 75 &296\\
       Subject 15   & 72 & 295 &72 &295\\
       Subject 16   & 74 & 293 &74 &293\\
       Subject 17   & 73 & 295 &73 &295\\
    \bottomrule
    \end{tabular}
    \caption{EEG-fMRI Data: Number of Trials per Subject}
    \label{tab:eeg_fmri_trial_}
\end{table}

\section{Parameter Selection}
\subsection{Multimodal Tensor Factorization}
The proposed model requires the selection of three different parameters, namely, $\gamma$, $\beta$, and rank $r$. To select these parameters, we closely follow best practices outlined in previous work on CMTF \citep{acar2011all}, ACMTF \citep{acar2014structure} and CCMTF \citep{mosayebi2020correlated}. First of all, one of these parameter can be set to 1 as a pivot, and following previous work, we set $\gamma=1$. The selection of rank $r$ is directly related to the selection of $\beta$. As $\beta$ enforces sparsity over the singular values, it directly minimizes the rank. With sufficiently large $r$, we can estimate the low-rank part through optimization. For the selection of $r$ in real data, we set $r=5$ following the work of \cite{mosayebi2020correlated} where it was shown through CORCONDIA tests \citep{kamstrup2013core} that $r=3$ is sufficiently large for oddball data. In the case of the simulation study, $r=5$ is again sufficiently large as the data were generated from rank $r=3$ factors. Finally, based on our empirical results and the results presented in \citep{mosayebi2020correlated} we set $\beta=0.001$ using k-fold cross validation. 

\subsection{C-STM}
The parameters in C-STM include kernel weights $w_{1}, w_{2}, w_{3}$ and regularization parameter $\lambda$ in the optimization. The weight parameters are selected based on the procedure in \citet{gonen2011multiple}, and $\lambda$ is selected using cross-validation. 

\bibliographystyle{apalike}
\bibliography{reference}

\end{document}